\documentclass[10pt,journal]{IEEEtran}
\usepackage{blindtext, graphicx}
\usepackage[binary-units=true]{siunitx}
\usepackage[nomessages]{fp}
\usepackage{verbatim}
\usepackage[super]{nth}
\usepackage{soul}
\usepackage{subfig}
\usepackage{multirow}
\usepackage{algpseudocode}
\usepackage{threeparttable}
\usepackage{booktabs}
\usepackage{footnote}
\usepackage{colortbl}
\usepackage{tabulary}
\usepackage[binary-units=true]{siunitx}

\definecolor{red}{rgb}{1,0,0}
\usepackage{todonotes}
\usepackage[euler]{textgreek}
\usepackage{subfiles}
\usepackage{amsmath}
\usepackage{float}
\newfloat{listing}{tbh}{lol}
\floatname{listing}{Listing}

\newcommand{\TP}{{\footnotesize \texttt{TP}}}
\newcommand{\uC}{MCU} 

\renewcommand{\baselinestretch}{0.93}
\newcommand{\bestefficiency} {21.6}

\usepackage[final]{changes}
\setdeletedmarkup{}

\ifCLASSOPTIONcompsoc
  \usepackage[nocompress]{cite}
\else
  \usepackage{cite}
\fi

\begin{document}

\title{XNOR Neural Engine: a Hardware Accelerator IP\\ for \bestefficiency{} fJ/op Binary Neural Network Inference}

%
%
%
%

\author{Francesco~Conti,~\IEEEmembership{Member,~IEEE,}
   	Pasquale~Davide~Schiavone,~\IEEEmembership{Student~Member,~IEEE,}
    and~Luca~Benini,~\IEEEmembership{Fellow,~IEEE}
\thanks{
This article will be presented in the International Conference on Hardware/Software Codesign and System Synthesis 2018 (CODES'18) and will appear as part of the ESWEEK-TCAD special issue.
This work was partially supported by Samsung under the GRO project "SCAlable Learning-in-place Processor".

F. Conti and L. Benini are with the Integrated Systems Laboratory, D-ITET, ETH Z\"urich, 8092 Z\"urich, Switzerland and with the Energy-Efficient Embedded Systems Laboratory, DEI, University of Bologna, 40126 Bologna, Italy. 
P. D. Schiavone is with the Integrated Systems Laboratory, D-ITET, ETH Z\"urich, 8092 Z\"urich, Switzerland.

E-mail: \texttt{\{fconti,pschiavo,lbenini\}@iis.ee.ethz.ch}.
\protect\\}
}


%
%

\markboth{IEEE Transactions on Computer-Aided Design of Integrated Circuits and Systems}%
{Conti \MakeLowercase{\textit{et al.}}, XNOR Neural Engine: a Hardware Accelerator IP for 21.6 fJ/op Binary Neural Network Inference}
%



\IEEEtitleabstractindextext{%
\begin{abstract}
Binary Neural Networks (BNNs) are promising to deliver accuracy comparable to conventional deep neural networks at a fraction of the cost in terms of memory and energy.
In this paper, we introduce the XNOR Neural Engine (XNE), a fully digital configurable hardware accelerator IP for BNNs, integrated within a microcontroller unit (\uC{}) equipped with an autonomous I/O subsystem and hybrid SRAM / standard cell memory.
The XNE is able to fully compute convolutional and dense layers in autonomy or in cooperation with the core in the \uC{} to realize more complex behaviors.
We show post-synthesis results in 65nm and 22nm technology for the XNE IP and post-layout results in 22nm for the full \uC{} indicating that this system can drop the energy cost per binary operation to \bestefficiency{}fJ per operation at 0.4V, and at the same time is flexible and performant enough to execute state-of-the-art BNN topologies such as ResNet-34 in less than 2.2mJ per frame at 8.9 fps.
\end{abstract}

}

\maketitle

\IEEEdisplaynontitleabstractindextext

%
\IEEEpeerreviewmaketitle

\section{Introduction}

\sloppypar
\IEEEPARstart{T}{oday}, \textit{deep learning} enables specialized cognition-inspired inference from collected data for a variety of different tasks such as computer vision~\cite{SzegedyRethinkingInceptionArchitecture2015}, voice recognition~\cite{ZhangEndtoEndSpeechRecognition2017}, big data analytics~\cite{ChenBigDataDeep2014}, financial forecasts~\cite{DixonImplementingDeepNeural2015}.
However, this technology could unleash an even higher impact on ordinary people's life if it was not limited by the constraints of data center computing, such as high latency and dependency on radio communications, with its privacy and dependability issues and hidden memory costs.
Low-power, embedded deep learning could potentially enable vastly more intelligent implantable biomedical devices~\cite{GreenspanGuestEditorialDeep2016}, completely autonomous nano-vehicles~\cite{LoquercioDroNetLearningFly2018} for surveillance and search\&rescue, cheap controllers that can be ``forgotten'' in environments such as buildings~\cite{ManicIntelligentBuildingsFuture2016}, roads, and agricultural fields.
As a consequence, there has been significant interest in the deployment of deep inference applications on microcontroller-scale devices~\cite{LaiCMSISNNEfficientNeural2018} and internet-of-things endnodes~\cite{ContiIoTEndpointSystemonChip2017}.
This essentially requires to fit the tens of billions of operations of a net such as ResNet-18~\cite{HeDeepResidualLearning2015} or Inception-v3/v4~\cite{SzegedyRethinkingInceptionArchitecture2015}\cite{SzegedyInceptionv4InceptionResNetImpact2016} on devices with a power budget of a few \SI{}{\milli\watt} costing less than 1\$ per device.

To meet these constraints, researchers have focused on reducing \textit{i}) the \textit{number of elementary operations}, with smaller DNNs~\cite{IandolaSqueezeNetAlexNetlevelaccuracy2016} and techniques to prune unnecessary parts of the network~\cite{HanLearningbothWeights2015}; \textit{ii}) the \textit{cost of an elementary compute operation}, by realizing more efficient software~\cite{LaiCMSISNNEfficientNeural2018} and hardware~\cite{HanEIEEfficientInference2016} and lowering the complexity of elementary operations~\cite{ZhouDoReFaNetTrainingLow2016}\cite{MoonsMinimumEnergyQuantized2017}; and \textit{iii}) the \textit{cost of data movement}, again by reducing the size of DNNs and taking advantage of locality whenever possible~\cite{11585_613492}.

An emerging trend to tackle \textit{ii}) and \textit{iii}) is that of fully binarizing both weights and activations in \textit{Binary Neural Networks} (BNNs)~\cite{CourbariauxBinarizedNeuralNetworks2016}\cite{RastegariXNORNetImageNetClassification2016}.
Their classification capabilities, together with the greatly reduced computational workload, represent a promising opportunity for integration in devices ``at the edge'', and even directly inside sensors~\cite{RusciDesignAutomationBinarized2017}.
Dropping the precision of weights and activations to a single bit enables the usage of simple XNOR operations in place of full-blown products, and greatly reduces the memory footprint of deep learning algorithms.

Software-based implementations of BNNs require special instructions for the popcount operation to be efficient and - more significantly - they require temporary storage of non-binary partial results either in the register file (with strong constraints on the final performance) or in memory (partially removing the advantage of binarization).
In this paper, we contribute the design of the \textit{XNOR Neural Engine} (XNE), a hardware accelerator IP for BNNs that is optimized for integration in a tiny microcontroller (\uC{}) system for edge computing applications.
While being very small, it allows to overcome the limitations of SW-based BNNs and execute fast binarized convolutional and dense neural network layers while storing all partial results in its internal optimized buffer.
We show that integrating the XNE within a \uC{} system leads to a flexible and usable accelerated system, which can reach peak efficiency of \SI{\bestefficiency{}}{\femto\joule} per operation but at the same time can be effectively used in real-world applications as it supports commonplace state-of-the-art BNNs such as ResNet-18 and ResNet-34 at reasonable frame rates ($>$8~fps) in less than 2.2 mJ per frame -- a third of a millionth of the energy stored in an AAA battery.
Finally, we show that even if binarization reduces the memory footprint and pressure with respect to standard DNNs, memory accesses and data transfers still constitute a significant part of the energy expense in the execution of real-world BNNs -- calling for more research at the algorithmic, architectural and technological level to further reduce this overhead.

\section{Related Works}
\label{sec:related}

\begin{table*}[tb]
\footnotesize
\centering
    \begin{tabulary}{\textwidth}{l C C}
        \textbf{Dataset / Network} & \textbf{Top-1 Acc.} & \textbf{CONV / FC weights} \\
        \toprule
        MNIST / fully connected BNN~\cite{CourbariauxBinarizedNeuralNetworks2016}    & 99.04~\%                    & - / \SI{1.19}{\mega\byte} \\
        SVHN / fully connected BNN~\cite{CourbariauxBinarizedNeuralNetworks2016}     & 97.47~\%                    &  \SI{139.7}{\kilo\byte} / \SI{641.3}{\kilo\byte} \\
        CIFAR-10 / fully connected BNN~\cite{CourbariauxBinarizedNeuralNetworks2016} & 89.95~\%                    &  \SI{558.4}{\kilo\byte} / \SI{1.13}{\mega\byte} \\
        ImageNet / ResNet-18 XNOR-Net~\cite{RastegariXNORNetImageNetClassification2016}   & 51.2~\%  & \SI{1.31}{\mega\byte} / \SI{2.99}{\mega\byte} \\
        ImageNet / ResNet-18 ABC-Net M=3,N=3~\cite{LinAccurateBinaryConvolutional2017}    & 61.0~\%  & \SI{3.93}{\mega\byte} / \SI{8.97}{\mega\byte} \\
        ImageNet / ResNet-18 ABC-Net M=5,N=5~\cite{LinAccurateBinaryConvolutional2017}    & 65.0~\%  & \SI{6.55}{\mega\byte} / \SI{14.95}{\mega\byte} \\
        ImageNet / ResNet-34 ABC-Net M=1,N=1~\cite{LinAccurateBinaryConvolutional2017}    & 52.4~\%  & \SI{2.51}{\mega\byte} / \SI{2.99}{\mega\byte} \\
        ImageNet / ResNet-34 ABC-Net M=3,N=3~\cite{LinAccurateBinaryConvolutional2017}    & 66.7~\%  & \SI{7.54}{\mega\byte} / \SI{8.97}{\mega\byte} \\
        ImageNet / ResNet-34 ABC-Net M=5,N=5~\cite{LinAccurateBinaryConvolutional2017}    & 68.4~\%  & \SI{12.57}{\mega\byte} / \SI{14.95}{\mega\byte} \\
        \bottomrule
    \end{tabulary}
    \caption{\added{BNNs proposed in literature, along with the related top-1 accuracy and weight memory footprint.}}
    \label{tab:soa_accuracy}
\end{table*}

The success of Deep Learning and, in particular convolutional neural networks, has triggered an exceptional amount of interest in hardware architects and designers who have tried to devise the most efficient way to deploy this powerful class of algorithms on embedded computing platforms.
Given the number of designs that have been published for CNNs, we will focus on a more direct comparison with accelerators that explicitly target a tradeoff between accuracy and energy or performance, keeping in mind that state-of-the-art accelerators for ``conventional'' fixed-point accelerators such as Orlando~\cite{Desoli9TOPSdeepconvolutional2017} are able to reach energy efficiencies in the order of a few Top/s/W.

The approaches used to reduce energy consumption in CNNs can be broadly categorized in two categories, sometimes applied simultaneously.
The first approach is to prune some calculations to save time and energy, while performing the rest of the computations in ``full precision''.
One of the simplest techniques is that employed by \textit{Envision}~\cite{MoonsEnergyefficientConvNetsapproximate2016} by applying Huffman compression to filters and activations, therefore saving a significant amount of energy in the transfer of data on- and off-chip.
A similar technique, enhanced with learning-based pruning of ``unused'' weights, has been also proposed by Han~et~al.~\cite{HanLearningbothWeights2015} and employed in the \textit{EIE}~\cite{HanEIEEfficientInference2016} architecture.
\textit{NullHop}~\cite{AimarNullHopFlexibleConvolutional2017} exploits activation sparsity to reduce the number of performed operations by a factor of 5-10$\times$ (for example, up to 84\% of input pixels are nil in several layers of ResNet-50).

The other popular approach is to drop the arithmetic precision of weights or activations, to minimize the energy spent in their computation.
Up to now, this approach has proven to be very popular on the algorithmic side: DoReFaNet~\cite{ZhouDoReFaNetTrainingLow2016}, BinaryConnect~\cite{CourbariauxBinaryConnectTrainingDeep2015}, BinaryNet~\cite{CourbariauxBinarizedNeuralNetworks2016} and XNOR-Net~\cite{RastegariXNORNetImageNetClassification2016} have been proposed as techniques to progressively reduce the precision of weights and activations by quantizing it to less than 8 bits or outright binarizing it, at the cost of retraining and loss of accuracy.
More recently, methods such as ABC-Net~\cite{LinAccurateBinaryConvolutional2017} and Incremental Network Quantization~\cite{ZhouIncrementalNetworkQuantization2016} have demonstrated that low-precision neural networks can be trained to an accuracy decreased $<5\%$ with respect to the full precision one.
\added{Table \ref{tab:soa_accuracy} lists some of the BNNs proposed in the state-of-the-art, along with their accuracy and memory footprint.}
Naturally, this approach lends itself well to being implemented in hardware.
The \textit{Fulmine} SoC~\cite{ContiIoTEndpointSystemonChip2017} includes a vectorial hardware accelerator capable of scaling the precision of weights from 16 bits down to 8 or 4 bits, gaining increased execution speed with similar power consumption.
\textit{Envision}~\cite{MoonsEnergyefficientConvNetsapproximate2016} goes much further: it employs dynamic voltage, frequency and accuracy scaling to tune the arithmetic precision of its computation, reaching up to 10~Top/s/W.
\textit{YodaNN}~\cite{AndriYodaNNArchitectureUltraLow2017} drops the precision of weights to a single bit by targeting binary-weight networks (activations use ``full'' 12-bit precision), and can reach up to 61~Top/s/W using standard cell memories to tune down the operating voltage.

To reach the highest possible efficiency, binary and ternary neural networks are perhaps most promising as they minimize the energy spent for each elementary operation, and also the amount of data transferred to/from memory, which is one of the biggest contributors to the ``real'' energy consumption.
One of the first architectures to exploit these peculiarities has been \textit{FINN}~\cite{UmurogluFINNFrameworkFast2017}, which is able to reach more than 200 Gop/s/W on a Xilinx FPGA, vastly outperforming the state-of-the-art for FPGA-based deep inference accelerators.
Recent efforts for the deployment of binary neural networks on silicon, such as \textit{BRein}~\cite{AndoBReinMemorySingleChip2017}, \textit{XNOR-POP}~\cite{JiangXNORPOPprocessinginmemoryarchitecture2017}, \textit{Conv-RAM}~\cite{BiswasConvRAMEnergyEfficientSRAM} and Khwa~et~al.~\cite{Khwa65nm4KbAlgorithmDependent} have mainly targeted in-memory computing, with energy efficiencies in the range 20-55~Top/s/W.
However, the advantage of this methodology is not yet clear, as more ``traditional'' ASICs such as \textit{UNPU}~\cite{LeeUNPU506TOPS} and \textit{XNORBIN}~\cite{BahouXNORBIN95TOp2018} can reach a similar level of efficiency of 50-100~Top/s/W.
Finally, mixed-signal approaches~\cite{BankmanAlwaysOn8mJ86} can reach 10$\times$ higher efficiency, with much steeper non-recurrent design and verification costs.

Our work in this paper tries to answer a related, but distinct question with respect to the presented state-of-the-art: how to design a BNN accelerator tightly integrated within a microcontroller (so that SW and HW can efficiently cooperate) -- and how to make so while taking into account the system level effects related to memory which inevitably impact real-world BNN topologies such as ResNet and Inception.
Therefore, we propose a design based on the tightly-coupled shared memory paradigm~\cite{conti_ultralowenergy_2015} and evaluate its integration in a simple, yet powerful, microcontroller system.

\section{Architecture}
\label{sec:archi}

\subsection{Binary Neural Networks primer}
In binary neural networks, inference can be mapped to a sequence of convolutional and densely connected layers of the form
\begin{equation}
    \mathbf{y}(k_{out}) = \mathrm{bin}_{\pm1}\left(\mathbf{b}_{k_{out}} + \sum_{k_{in}} \Big( \mathbf{W}(k_{out},k_{in}) \otimes \mathbf{x}(k_{in}) \Big)\right)
    \label{eq:conv_base}
\end{equation}
where $\mathbf{W}$, $\mathbf{x}$, $\mathbf{y}$ are the binarized  ($\in \pm 1$) weight, input and output tensors respectively; $\mathbf{b}$ is a real-valued bias; $\otimes$ is the cross-correlation operation for convolutional layers and a normal product for densely connected ones.
$\mathrm{bin}_{\pm1}(\cdot)$ combines batch normalization for inference with binarization of the integer-valued output of the sum in Equation~\ref{eq:bin_base}:
\begin{equation}
    \mathrm{bin}_{\pm1}(t) = \mathrm{sign} \left( \gamma\frac{t - \mu}{\sigma} + \beta \right)
    \label{eq:bin_base}
\end{equation}
where $\beta$, $\gamma$, $\mu$, $\sigma$ are the learned parameters of batch normalization.

A more convenient representation of the BNN layer can be obtained by mapping elements of value $+1$ to 1-valued bits and those of value $-1$ to 0-valued bits, and moving the bias inside the binarization function.
Equation~\ref{eq:bin_base} can be reorganized into
\begin{equation}
    \mathrm{bin}_{0,1}(t) = 
    \begin{cases} 
        \text{1 if } t \geq -{\kappa}/{\lambda} \doteq \tau, \text{ else 0}\;\text{(when $\lambda > 0$)} \\
        \text{1 if } t \leq -{\kappa}/{\lambda} \doteq \tau, \text{ else 0}\;\text{(when $\lambda < 0$)} \\
   \end{cases}
    \label{eq:bin_hw_tau}
\end{equation}
where $\lambda \doteq \gamma/\sigma$, $\kappa \doteq \beta+\gamma/\sigma (b-\mu)$, and $\tau \doteq -\kappa/\lambda$ is a threshold defined for convenience in Section~\ref{sec:archi_datapath}.
Multiplications in Equation~\ref{eq:conv_base} can be replaced with XNOR operations, and sums with popcounting (i.e., counting the number of bits set to 1):
\begin{equation}
    \mathbf{y}(k_{out}) = \mathrm{bin}_{0,1}\left( \sum_{k_{in}} \Big( \mathbf{W}(k_{out},k_{in}) \otimes \mathbf{x}(k_{in}) \Big)\right)
    \label{eq:conv_hw}
\end{equation}

\subsection{XNE operating principles}
\label{sec:archi_principle}
  
\begin{listing}[tb]
\includegraphics{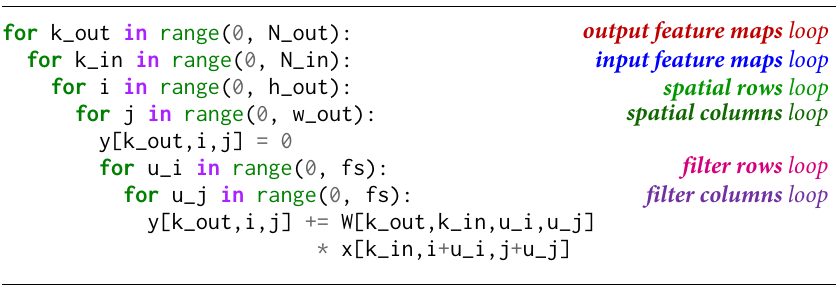}
\caption{Baseline loops of a BNN convolutional layer\protect\footnotemark.} 
\label{lst:std_conv}
\end{listing}
\footnotetext{\label{footnote:1}The \texttt{*} and \texttt{+=} operators indicate XNOR and popcount-accumulation respectively.
}


The XNOR Neural Engine we propose in this work has been designed to be able to execute both binarized convolutional and binarized dense layers.
Convolutional layers consist of six nested loops on output feature maps, input feature maps, two output spatial dimensions, and two filter spatial dimensions; Listing~\ref{lst:std_conv} shows a na\"ive implementation of a convolutional layer in Python pseudo-code.
Densely connected layers can be considered as a limit case of the convolutional layer for a $1\times 1$ filter on a single pixel in the spatial dimensions, i.e. {\footnotesize \texttt{h\_out}=\texttt{w\_out}=\texttt{fs}=1}.

In modern topologies~\cite{HeDeepResidualLearning2015}\cite{SzegedyInceptionv4InceptionResNetImpact2016}, deeper convolutional layers have {\footnotesize \texttt{N\_out},\texttt{N\_in}$>$\texttt{h\_out},\texttt{w\_out}}; in other words, layers become ``less convolutional'' and more similar to densely connected layers.
This leads towards choosing an architecture where pixel- or feature map-level parallelism is favored over filter-level parallelism (contrary to designs based on sliding windows).
This is particularly true for BNNs, where energy efficiency can be attained only by operating on tens/hundreds of binary pixels in parallel -- which cannot be done with filter-level parallelism on deeper layers.

A second fundamental consideration is that, since intermediate values of the \textit{popcount} operation are integer, it is highly preferable to perform the operation of Equation~\ref{eq:conv_hw} without storing them in memory.
In other words, the accelerator has to be weight- and output-stationary~\cite{UmurogluFINNFrameworkFast2017} or input- and output- stationary.
In the remainder on this paper, we focus exclusively on the latter case, although the XNE can arguably be used in both modes by swapping the roles of weights and inputs.
%
%

We designed the XNE around a lean hardware engine focused on the execution of the feature loops of Listing~\ref{lst:std_conv}.
We execute these as hardwired inner loops, operating in principle on a fixed-sized input tiles in a fixed number of cycles\footnote{The XNE can actually be configured to operate on smaller tiles when it is necessary, with a proportional decrease in loop latency.}.
A design-time \textit{throughput parameter} (\TP{}) is used to define the size of each tile, which is also the number of simultaneous XNOR operations the datapath can execute per cycle; every \TP{} cycles, the accelerator consumes one set of \TP{} input binary pixels and \TP{} sets of \TP{} binary weights to produce one set of \TP{} output pixels.

\begin{listing}[tb]
\includegraphics{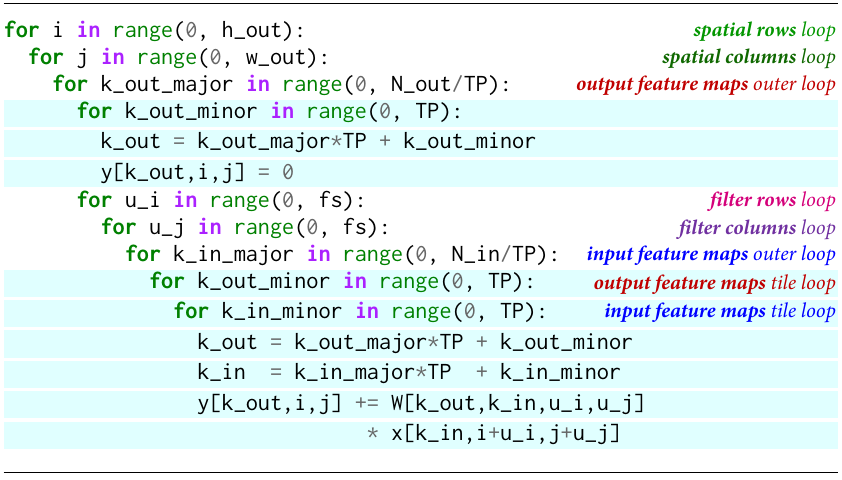}
\caption{Reordered DNN layer loops; the innermost loops (highlighted in light blue) are hardwired in the XNE engine, while the others can be implemented in the XNE microcode. Remainder loops are left out for simplicity.}
\label{lst:xne_conv}
\end{listing}

\replaced{
Listing~\ref{lst:xne_conv} shows how the convolutional layer is reorganized: \textit{i}) the loops are reordered, bringing spatial loops to the outermost position, feature-map loops to the innermost position and filter loops in the middle; \textit{ii}) the two feature loops are tiled and therefore split in a \textit{tile} loop (cycling on a tile of \TP{} iterations) and an \textit{outer} loop (cycling on {\footnotesize \texttt{nif/TP}} or {\footnotesize \texttt{nof/TP}} tiles); \textit{ii}) the output feature maps outer loop is moved outwards with respect to the filter loops.
}
{
Listing~\ref{lst:xne_conv} shows how the convolutional layer is reorganized by dividing the two feature loops in an inner part (cycling on a tile of \TP{} iterations) and an outer part (cycling on {\footnotesize \texttt{nif/TP}} or {\footnotesize \texttt{nof/TP}} tiles).
}
If {\footnotesize \texttt{nif}} and/or {\footnotesize \texttt{nof}} are not whole multiples of \TP{}, ``remainder'' iterations have to be executed; these are left out of the listing for the sake of brevity.
The innermost loops, which are shown highlighted in blue, are hardwired in the engine datapath as previously introduced and fully explained in Section \ref{sec:archi_datapath}, which details the datapath micro-architecture.
The outermost loops, instead, are implemented by means of a tiny microcode processor embedded in the XNOR Neural Engine, as detailed in Section \ref{sec:archi_ucode}.

\subsection{Accelerator architecture}
\begin{figure}
    \centering
    \includegraphics[clip, trim= 0pt 0pt 0pt 0pt, width=0.48\textwidth]{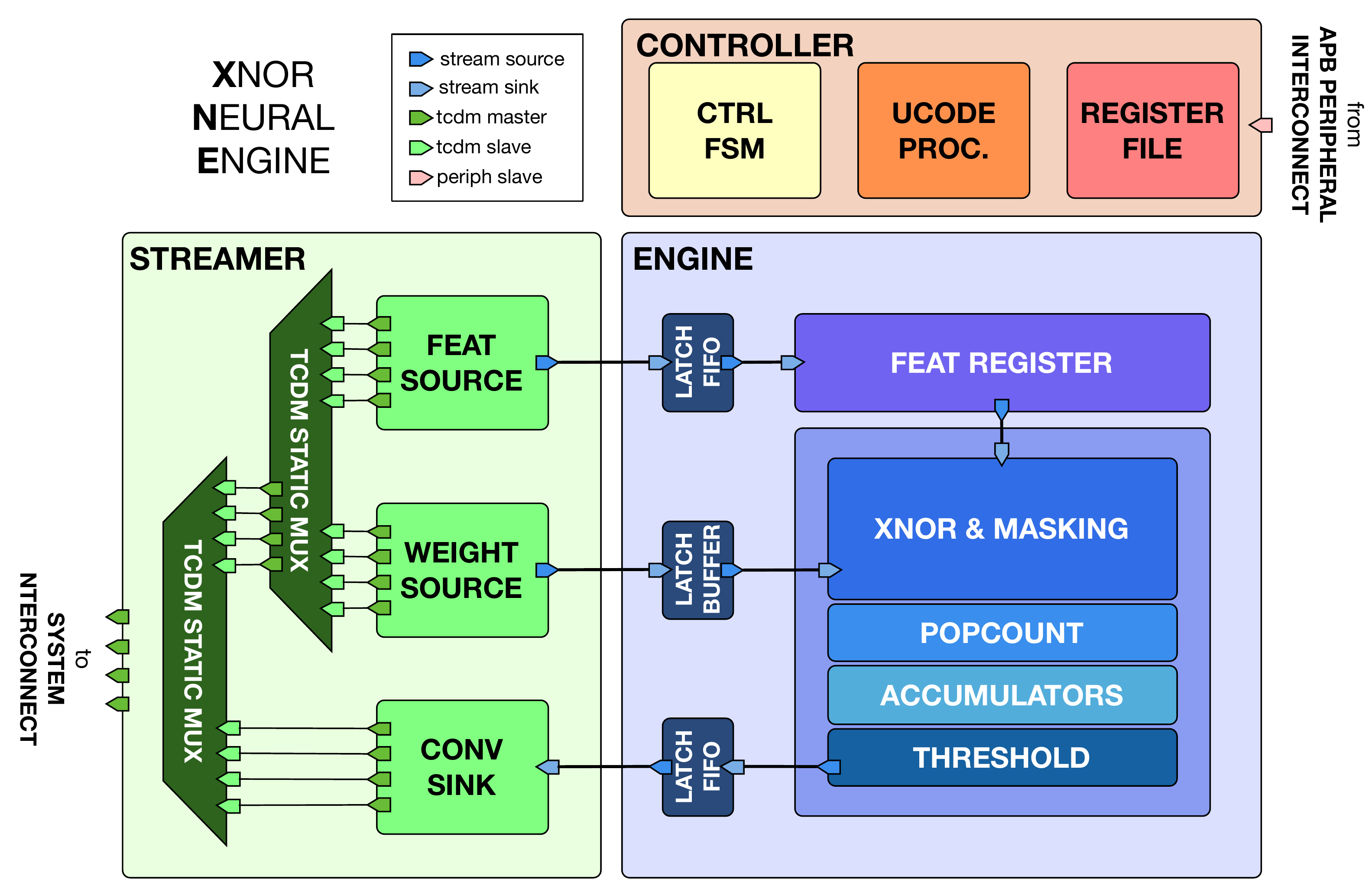}
    \caption{XNOR Neural Engine overall architecture for \TP{}=128.}
    \label{fig:xne_archi}
\end{figure}

Figure~\ref{fig:xne_archi} shows the overall architecture of the XNE.
The \textit{controller}, which can be targeted in a memory-mapped fashion via a target port using the AMBA APB protocol, is responsible of coordinating the overall operation of the accelerator.
It contains a latch-based register file, a central controller finite-state machine (FSM), and a \textit{microcode processor} (further detailed in Section \ref{sec:archi_ucode}) that is responsible of implementing the outer loops of Listing~\ref{lst:xne_conv}.
The \textit{engine} contains the streaming datapath, which executes the inner loop operation of Listing~\ref{lst:xne_conv}.
It operates on streams that use a simple valid-ready handshake similar to that used by AXI4-Stream~\cite{amba_axi4stream}.
Finally, the \textit{streamer} acts as a transactor between the streaming domain used by the internal engine and the memory system connected to the accelerator.
It is capable of transforming streams of width multiple of 32 bits into byte-aligned accesses to the cluster shared memory, and vice versa.

\begin{figure}
    \centering
    \includegraphics[clip, trim= 0pt 0pt 0pt 0pt, width=0.48\textwidth]{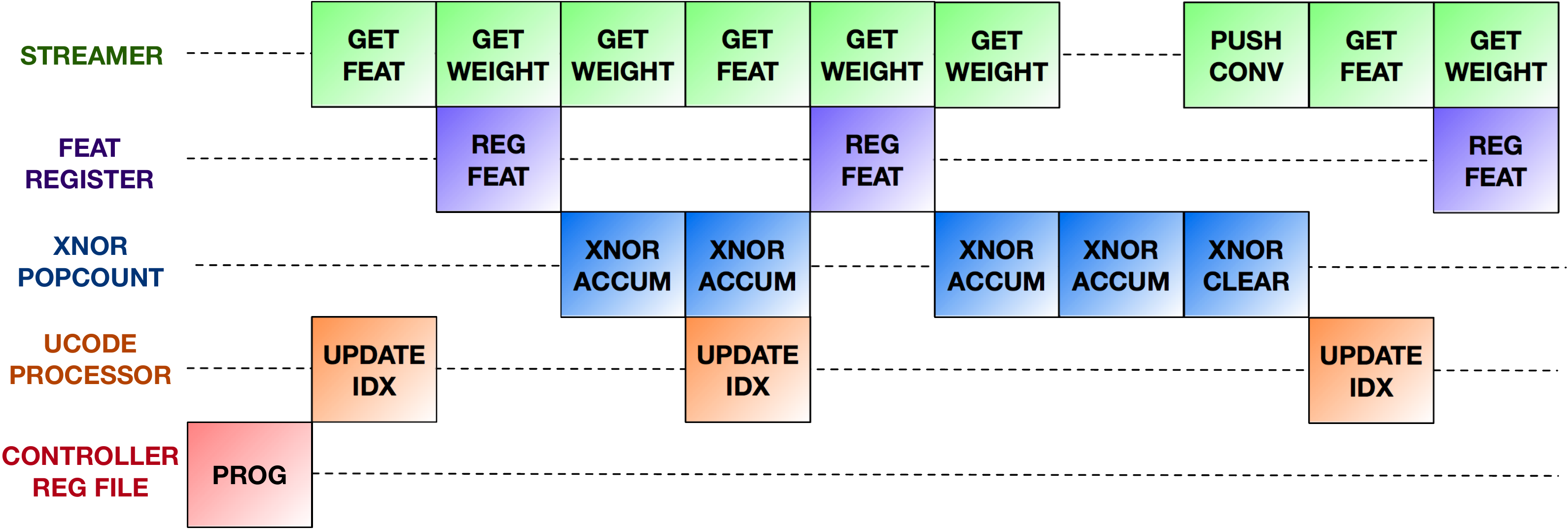}
    \caption{Example of XNE operation divided in its main phases.}
    \label{fig:xne_timing}
\end{figure}

Figure~\ref{fig:xne_timing} shows a high-level view of how the XNE operates.
The controller register file is first programmed with the DNN layer parameters (e.g. {\footnotesize\texttt{nif}}, {\footnotesize\texttt{nof}}, {\footnotesize\texttt{fs}}, etc.) and with the microcode byte code.
The central controller FSM then orchestrates the operation of the XNE, which is divided in three phases: \textsc{feature loading}, \textsc{accumulation}, \textsc{thresholding/binarization}.
In the \textsc{feature loading} phase, the \textit{i}-th feature \TP{}-vector is loaded from the streamer, while at the same time the microcode processor starts updating the indeces used to access the next one.
In the \textsc{accumulation}, for \TP{} iterations a new weight \TP{}-vector is loaded and multiplied by the feature vector, and the result is saved in an accumulator.
In the \textsc{thresholding and binarization} phase, \TP{} threshold values are loaded from memory and used to perform the binarization, then the binarized outputs are streamed out of the accelerator.
These three phases are repeated as many times as necessary to implement the full loop of Listing~\ref{lst:xne_conv}.

\subsubsection{Interface modules}
\label{sec:archi_interface}

The interface that the XNE exposes follows the paradigm of shared-memory, tightly coupled Hardware Processing Engines~\cite{conti_ultralowenergy_2015}.
The XNE has a single APB target port, which allows memory mapped control of the XNE and access to its register file, and $\TP{}/32$ master ports (each 32 bits wide) enabling access to the shared memory system via word-aligned memory accesses.
Finally, a single \textit{event} wire is used to signal the end of the XNE computation to the rest of the system.

The \textit{controller} module, which is the direct target of the slave port, consists of the memory-mapped register file, a finite-state machine used to implement the main XNE operation phases as shown in Figure~\ref{fig:xne_timing}, and a microcode processor to implement the loops in Listing~\ref{lst:xne_conv} (as described in Section \ref{sec:archi_ucode}).
The memory-mapped register file uses standard cell memories implemented with latches to save area and power with respect to conventional flip-flops.
It includes two sets of registers: \textit{generic} ones, used to host parameters that are assumed to be static between the execution of multiple jobs, and \textit{job-dependent} ones, for parameters that normally change at every new job (such as base pointers).
The latter set of registers is duplicated so that one new job can be offloaded from the controlling processor to the XNE even while it is still working on the current one.

The \textit{streamer} module contains the blocks necessary to move data in and out of the accelerator through its master ports, and transform the memory accesses into coherent streams to feed the accelerator inner engine\footnotemark.
\footnotetext{
	Controller and streamer IPs are available as open-source at \texttt{github.com/pulp-platform/hwpe-ctrl} and \texttt{github.com/pulp-platform/hwpe-stream} respectively.
}
\replaced
{These are organized in separate hardware modules, two \textit{sources} for incoming streams (one for weights/thresholds, one for input activations) and one \textit{sink} for the outgoing one (output activations).
Both the two sources and the sink include an own address generation block to start the transaction in memory and a realigner to transform vectors that start from a non-word-aligned base into well-formed streams, without assuming that the memory system outside of the accelerator can natively support misaligned accesses.
The memory accesses produced by the source and sink modules are mixed by two static mux/demux blocks; the controller FSM ensures that only one is active at any given cycle and that no transactions are lost.
}
{
To this end, each logical stream (one for weights and thresholds, one for input activations, one for output activations) has its own address generation block to start the appropriate memory transaction.
The streamer also includes realigners, to be able to transform vectors that start from a non-word-aligned base into well-formed streams, without assuming that the memory system outside of the accelerator can natively support misaligned accesses.
}

\subsubsection{Microcode processor}
\label{sec:archi_ucode}


\begin{listing}[tb]
\includegraphics{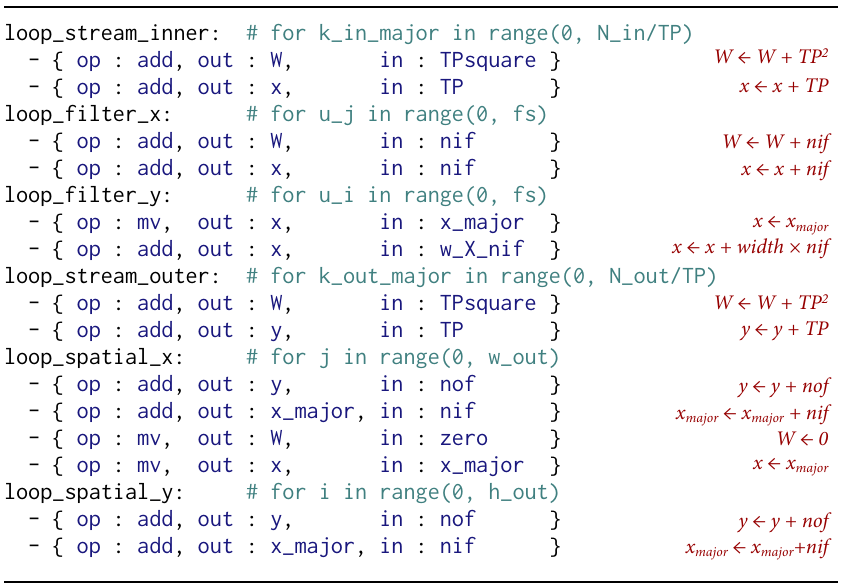}
%
\caption{Microcode specification for the six loops shown in Listing~\ref{lst:xne_conv}. {\footnotesize\texttt{W}}, {\footnotesize\texttt{x}}, {\footnotesize\texttt{y}} and {\footnotesize\texttt{x\_major}} are mnemonics for the four R/W registers; {\footnotesize\texttt{TPsquare}}, {\footnotesize\texttt{TP}}, {\footnotesize\texttt{nif}}, {\footnotesize\texttt{nof}}, {\footnotesize\texttt{w\_X\_nif}}, {\footnotesize\texttt{ow\_X\_nof}}, {\footnotesize\texttt{zero}} are mnemonics for the R/O registers used in this implementation.}
\label{lst:xne_ucode}
\end{listing}

\sloppypar
Instead of relying on an external processor to compute the new offsets for memory access, to iterate the inner loop execution, and to maintain overall state, the XNE can use a small internal \textit{microcode processor} to implement the six nested outer loops shown in Listing~\ref{lst:xne_conv}.
The microcode processor has four R/W registers, used to compute the {\footnotesize\texttt{i}},{\footnotesize\texttt{j}}, {\footnotesize\texttt{k\_out\_major}}, {\footnotesize\texttt{u\_i}},  {\footnotesize\texttt{u\_j}}, {\footnotesize\texttt{k\_in\_major}} indeces of Listing~\ref{lst:xne_conv}; and can access sixteen R/O registers.
The latter are used to store loop ranges and iteration values, coming from the register file directly or indirectly, i.e. computed from register file values using simple sequential multipliers to minimize hardware overhead.

The microcode processor uses a custom tiny ISA with two ``imperative'' instructions, \textit{ADD} (add/accumulate) and \textit{MV} (move).
They use one of the R/W registers as output and one R/O or R/W register as input; the \textit{ADD} instruction implicitly uses the output register as a second input.
The microcrode ISA also includes one ``declarative'' \textit{LOOP} instruction, containing the iteration range of each loop and the base address and number of micro-instructions associated to it.
The hardware implementation of this ISA is a single-stage execution pipeline controlled by four simple finite-state machines operating in lockstep; they compute the address of the next micro-instruction to execute, its index within the current loop, the next iteration index of the current loop, and the next loop to be taken into account.

The microcode associated to the functionality presented in Listing~\ref{lst:xne_conv} (six loops) occupies 28B in total (22B for the imperative part, 6B for the declarative one) which are mapped directly within the XNE register file.
The final microcode, which is specified in a relatively high-level fashion by means of a description in the YAML markup language, can be seen in  Listing~\ref{lst:xne_ucode}.
This description can be compiled into a bitstream using a simple Python script and added to the preamble of an application; the microcode is stored in the ``generic'' section of the register file and is kept between consecutive jobs unless explicitly changed.

\subsubsection{Datapath micro-architecture}
\label{sec:archi_datapath}

\begin{figure}
    \centering
    \includegraphics[clip, trim= 0pt 0pt 0pt 0pt, width=0.45\textwidth]{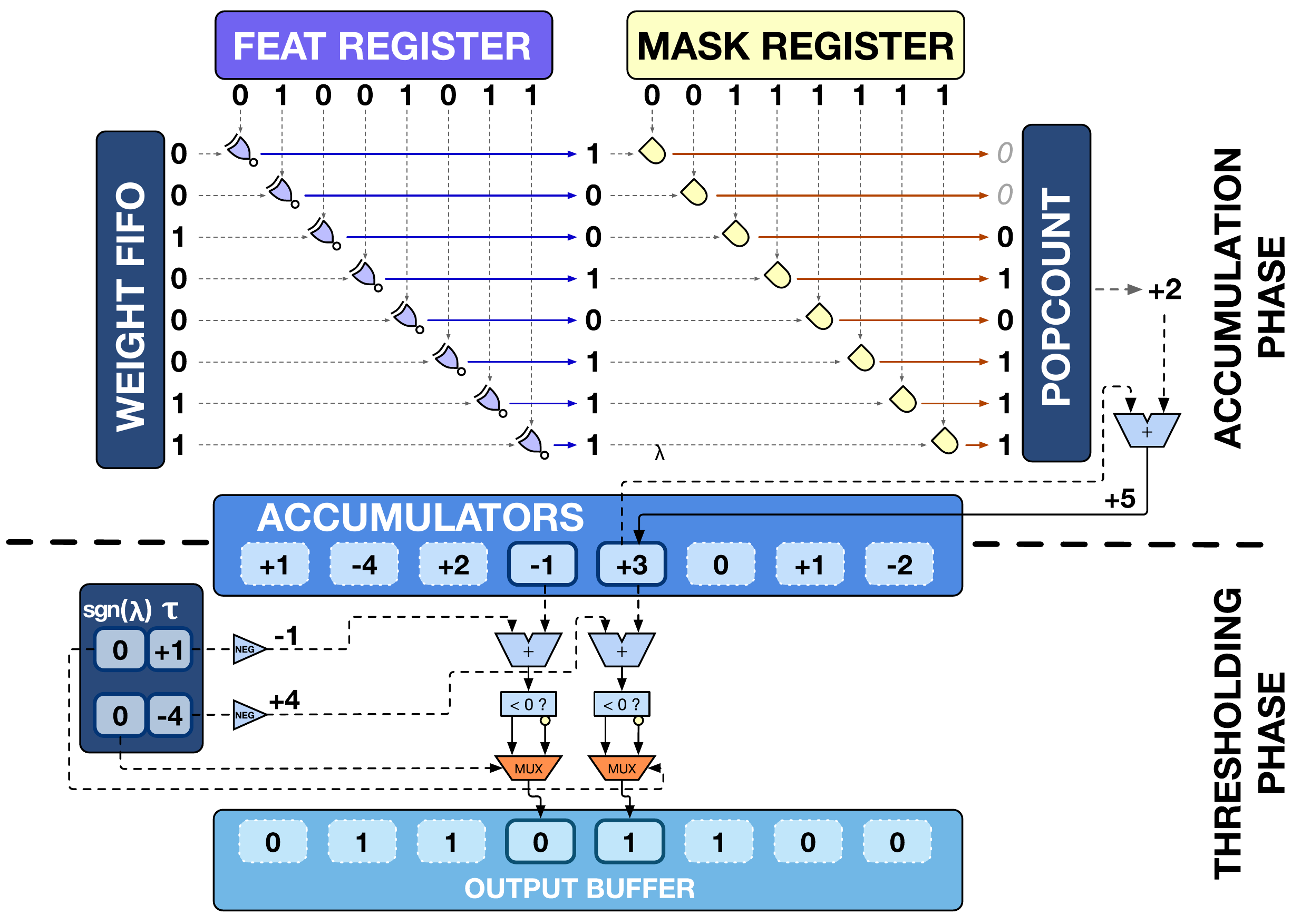}
    \caption{XNE datapath for XNOR, popcounting, accumulation and thresholding (\TP{}=8).}
    \label{fig:xne_datapath}
\end{figure}

The XNE datapath is composed by the blocks responsible of performing vector binary multiply (realized by means of XNOR gates), accumulation (within a latch-based register file) and thresholding to determine normalized binary outputs.
\added{The datapath is fed with the weight/threshold and the input activation streams coming from the streamer sources through two-element FIFOs; it produces an output activation stream into a decoupling two-element FIFO, which on turn is connected with the streamer sink.}
Figure~\ref{fig:xne_datapath} illustrates the structure of the datapath in a case where \TP{} is 8.
The input feature \TP{}-vector is stored in a \textit{feature register} to be reused for $\min$(\TP{},{\footnotesize\texttt{N\_out}}) cycles (one for each accumulator used).
Once an output feature vector has been produced by the XNE datapath, it is completely computed and never used again.
With the microcoding strategy proposed in Listing~\ref{lst:xne_ucode}, a single input feature vector has to be reloaded {\footnotesize \texttt{fs}}${}^2$ times, and afterwards it is completely consumed.

The weight \TP{}-vector stream produced by the streamer is decoupled from the main datapath by means of a four-element FIFO queue; at each cycle in the main binary convolution execution stage, the feature vector is ``multiplied'' with the weight stream by means of \TP{} XNOR gates, producing the binary contributions of all \TP{} input feature elements to a single output feature element.
These contributions are masked by means of an array of AND gates to allow the XNE to work even when the number of input features is smaller than \TP{}.
A combinational reduction tree is used to perform the \textit{popcount} operation, i.e. to count the number of 1's in the unmasked part of the product vector.
The output is accumulated with the current state of an accumulator register; there are in total \TP{} accumulators, one for each output computed in a full accumulation cycle.
Accumulated values are computed with 16 bit precision and saturated arithmetic.

To implement the binarization function of Equation~\ref{eq:bin_hw_tau}, the value stored in the accumulators is binarized after a thresholding phase, which encapsulates also batch normalization.
The binarization thresholds are stored in a vector of \TP{} bytes, and loaded only when the accumulated output activations are ready to be streamed out.
Each byte is composed of 7 bits (one for sign, six for mantissa) representing $\tau$, plus 1 bit used to represent $\mathrm{sign}(\lambda)$ (used to decide the sign of the comparison).
The 7-bit $\tau$ is left-shifted of a configurable amount of bits \added{$S_\mathrm{\tau}$}, to enable the comparison with the 16-bit accumulators.
The output of the thresholding phase is saved in a FIFO buffer, from which it is sent to the streamer module (see Figure~\ref{fig:xne_archi}) so that it can be stored in the shared memory.

\subsubsection{\added{Impact of accumulator and threshold truncation}}
\label{sec:quantization_impact}
\added{
According to our experiments, the impact of truncating accumulators (to 16 bits) and thresholds (to 7 bits) is very small.
Errors due to accumulator truncation can happen only on bigger layers than what is found in most BNN topologies (e.g., even a layer with {\footnotesize \texttt{nif}}=1024, {\footnotesize \texttt{fs}}=5 does not have enough accumulations per output pixel to hit the accumulator dynamic range), and only in consequence of unlikely imbalances between 0's and 1's; saturation provides a mitigation mechanism for many of these cases.
}

\added{
For what concerns the truncation of batch-normalization thresholds to 7 bits, if a shift $S_{\tau}>0$ is being used, a super-set of the accumulator values that could be affected (i.e. that could be binarized incorrectly) is given by the worst-case error interval $\left[\tau\pm{2^{S_{\tau}-1}}\right]$.
The probability that accumulator values reside within this interval (i.e., they are near the threshold between providing a +1 or -1) depends on the layer size and the training methodology, as well as the actual input of the BNN.
In our experiments of Section~\ref{sec:results_energy_accuracy} using the training method of Courbariaux~et~al.~\cite{CourbariauxBinarizedNeuralNetworks2016}, we did not observe any accuracy degradation with $S_{\tau}$ values (between 0 and 2) adequate to represent all the dynamic range of the thresholds.
}


\section{Experimental Results}
\label{sec:results}

In this section, we evaluate the energy and area efficiency of the proposed XNE accelerator design taken ``standalone'' with several choices of the \TP{} parameter; then we showcase and evaluate a full microcontroller system augmented with the XNE accelerator.

\subsection{Standalone XNE}
\label{sec:results_sweep}

\begin{figure*}[t]
    \centering
    \includegraphics[clip, trim= 0pt 0pt 0pt 0pt, width=0.95\textwidth]{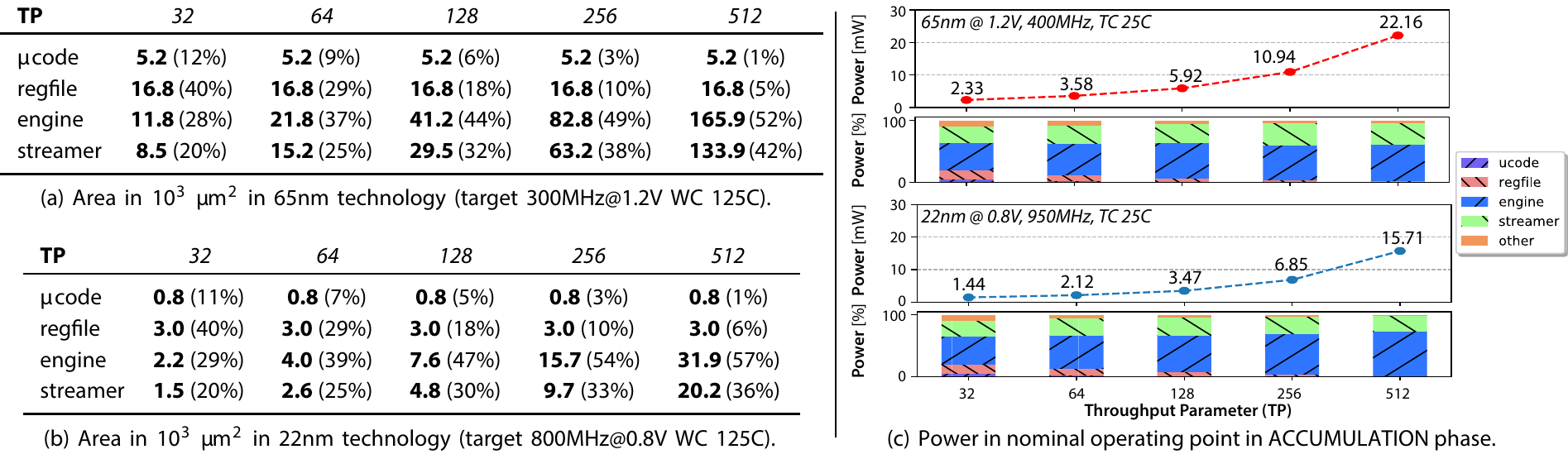}
    \caption{Stand-alone XNE results in terms of area and power in nominal operating conditions for the two target technologies.}
    \label{fig:xne_power_all}
\end{figure*}


\begin{figure*}[tb]
    \centering
    \subfloat[\uC{} architecture. \label{fig:quentin_archi_archi}]{
        \includegraphics[clip, width=0.40\textwidth]{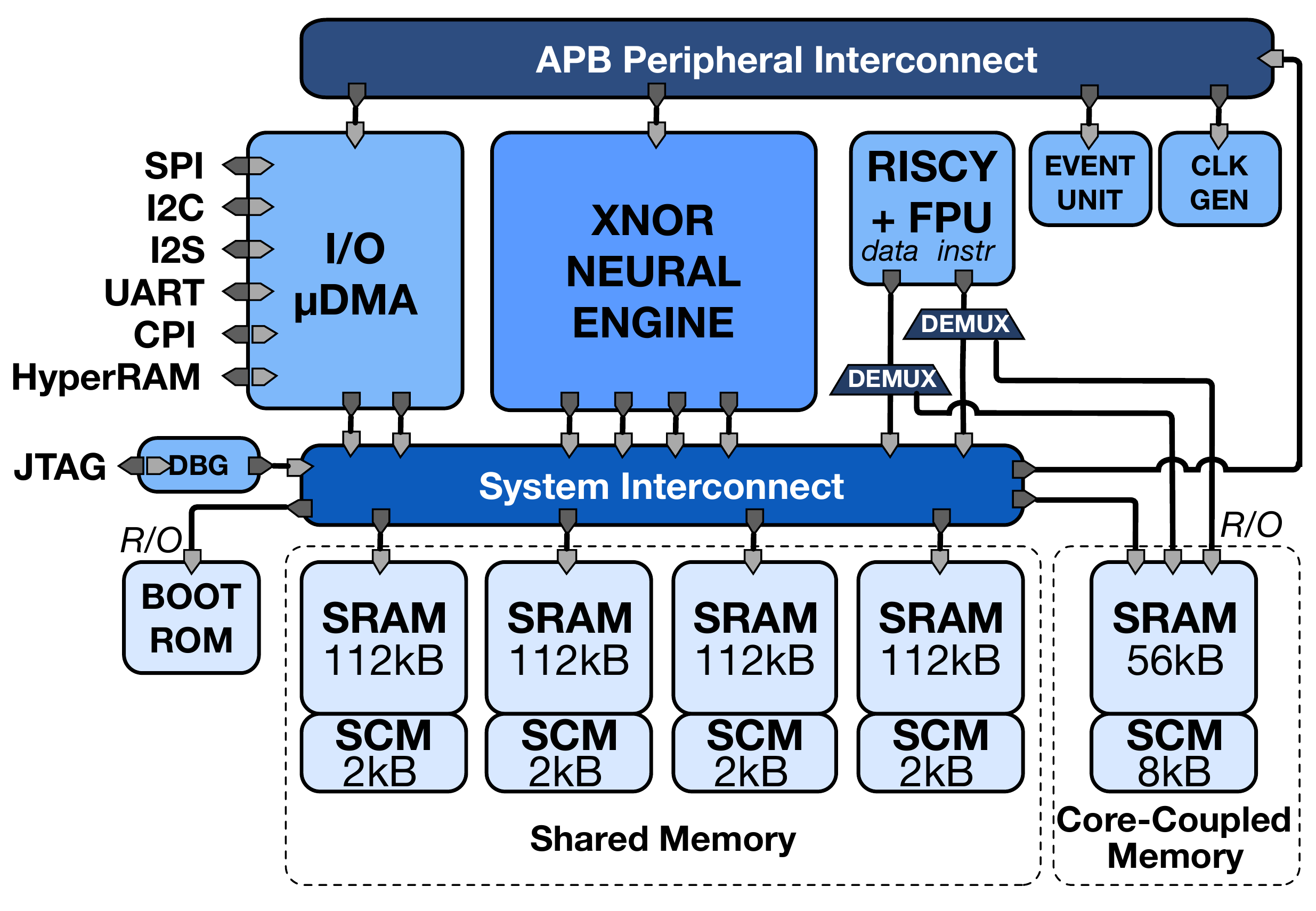}
    }
    \qquad\qquad\quad
    \subfloat[\uC{} floorplan in 22nm. \label{fig:quentin_floorplan}]{
        \includegraphics[clip, width=0.36\textwidth]{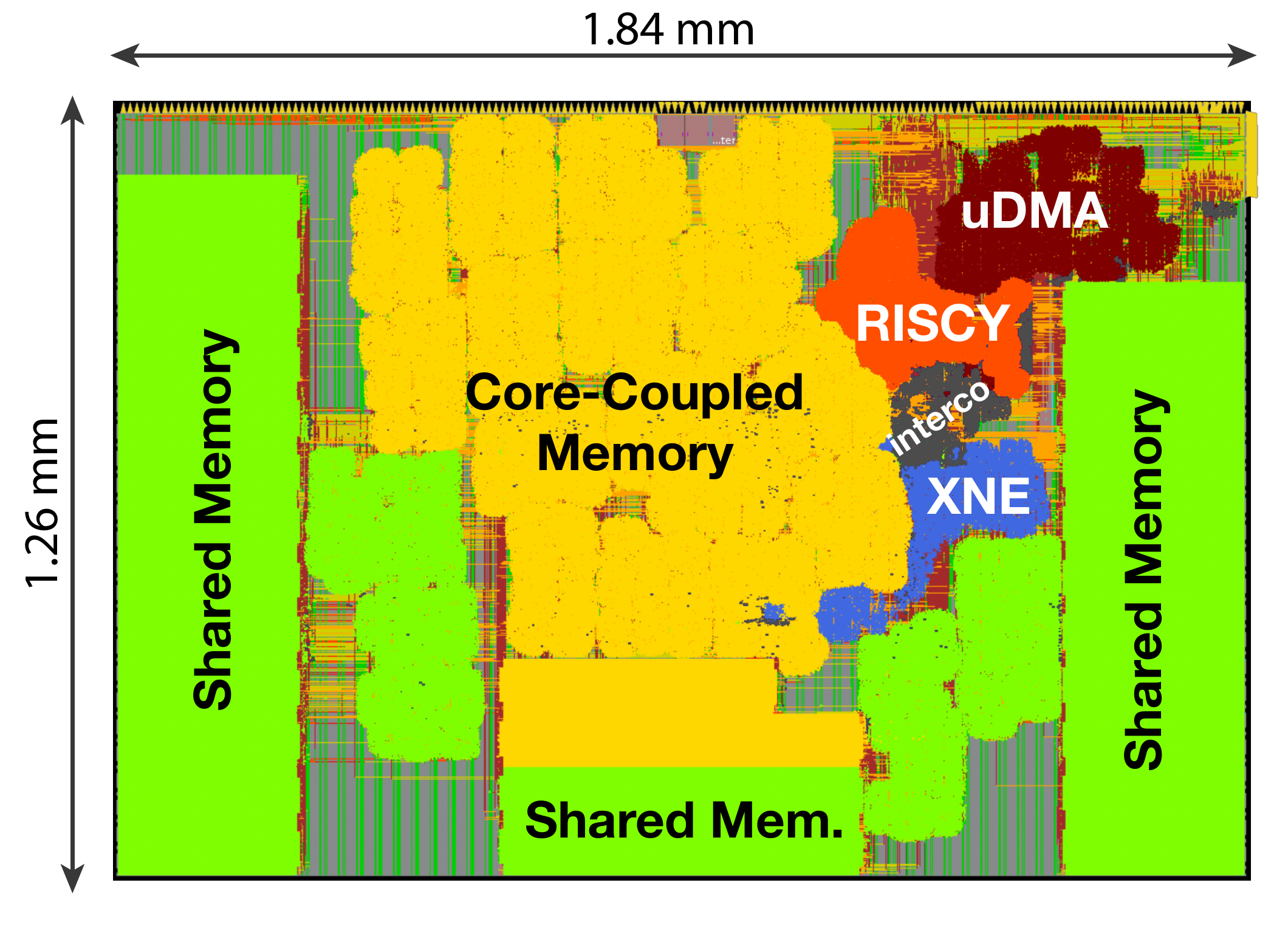}
    }
    \caption{Architecture of the microcontroller system (\uC{}) and its layout in 22nm technology.}
    \label{fig:quentin_archi}
\end{figure*}

The main architectural parameter of the XNE, the \textit{throughput parameter} \TP{}, can be used to choose the amount of hardware parallelism exploited by the accelerator, and the related required number of master ports on the memory side.
In this section, we make a first evaluation on how changing this parameter can influence the area and power of the accelerator.
We implemented the XNE in synthesizable SystemVerilog~HDL using \TP{} as a design-time parameter, sweeping from \TP{}=32 to \TP{}=512 in geometric progression.

%
The various versions of the XNE were synthesized using Synopsys Design Compiler 2017.09 targeting \SI{300}{\mega\hertz}@0.59V, 125C and \SI{800}{\mega\hertz}@1.08V, 125C in 65nm and 22nm, respectively (in worst case).
Afterwards, we performed a place~\&~route run of the block using Cadence Innovus 16.10.
We targeted 65\% utilization on a square area; as the XNE is synthesized stand-alone instead of in coupling with a multi-banked memory, this P\&R does not accurately model all effects present when deploying an XNE in a real platform.
However, it enables vastly more accurate power prediction with respect to post-synthesis results after clock tree synthesis and the extraction of wiring parasitics.
Moreover, the 65\% utilization target is conservative enough so that it is possible to check that the XNE does not introduce congestion when routed on a more realistic design
For power estimation, performed with Synopsys PrimeTime PX 2016.12, we used activity dumps from post-layout simulation and we targeted the typical corner.
After P\&R, all XNEs are able to work at up to \SI{400}{\mega\hertz}@1.25V, 25C (in 65nm) / \SI{950}{\mega\hertz}@0.72V, 25C (22nm) in the typical corner.
%


In Figure~\ref{fig:xne_power_all}, we report the area of the synthesized XNE with the 65nm and 22nm libraries; the Table shows that the fixed costs of the microcode processor and register file are progressively absorbed as the size of the engine and streamer increase near-linearly with \TP{}.
Figure~\ref{fig:xne_power_all} also reports power estimation results in nominal operating conditions from the various versions of the XNE (in the active \textsc{accumulation} phase), shows similar scaling, with the engine and streamer modules being responsible for most of the power consumed by the XNE.
The latter point indicates that, as expected, the XNE shows a high internal architectural efficiency.

\subsection{XNE in a \uC{} System}
\label{sec:results_uC}

\begin{figure*}[tb]
\footnotesize
\centering
    \subfloat[Performance in Gop/s. \label{fig:perf}]{
        \includegraphics[clip, width=0.28\textwidth]{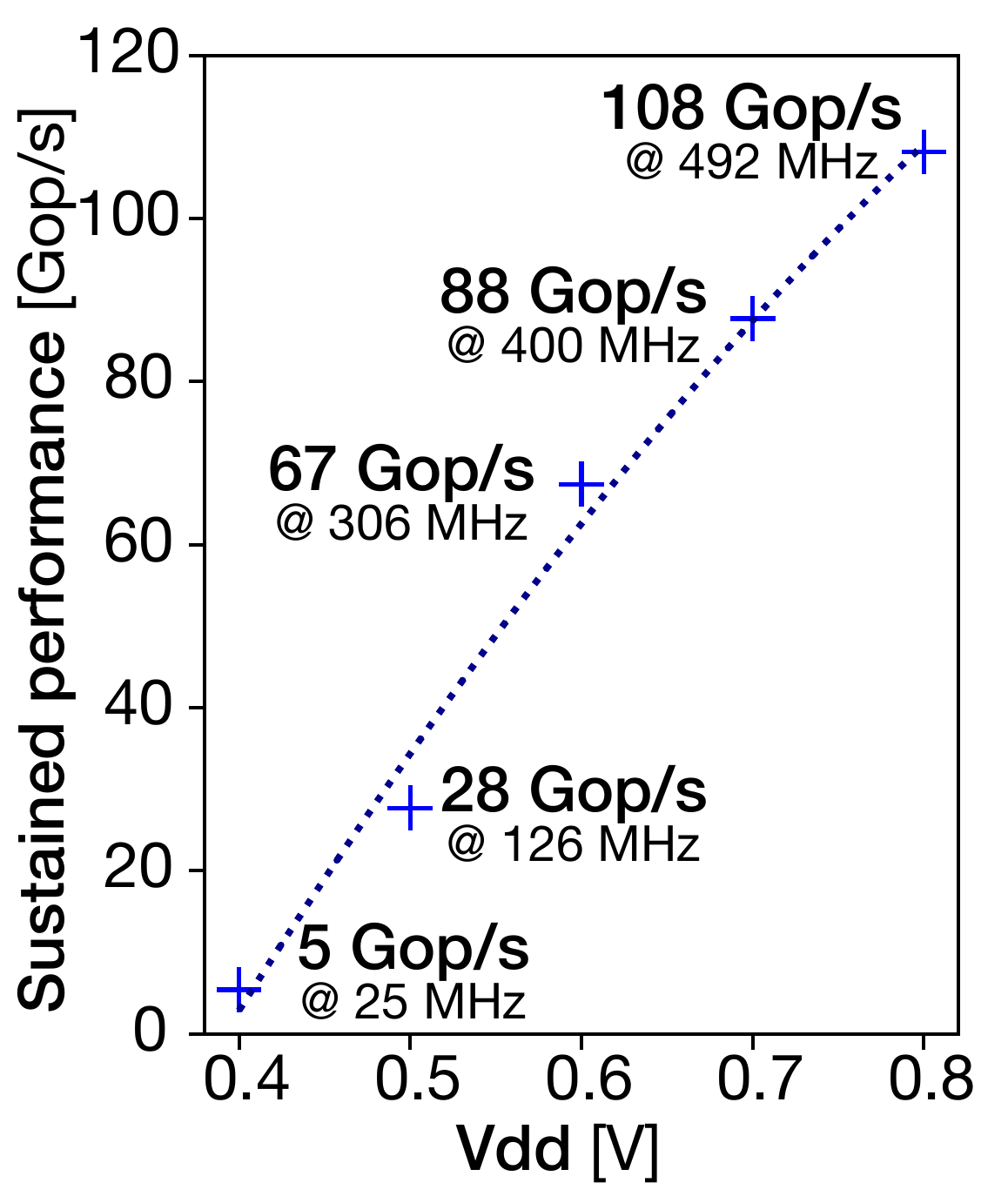}
    }
    \qquad\qquad\quad
    \subfloat[\uC{}-level energy efficiency in fJ/op. \label{fig:efficiency}]{
        \includegraphics[clip, width=0.48\textwidth]{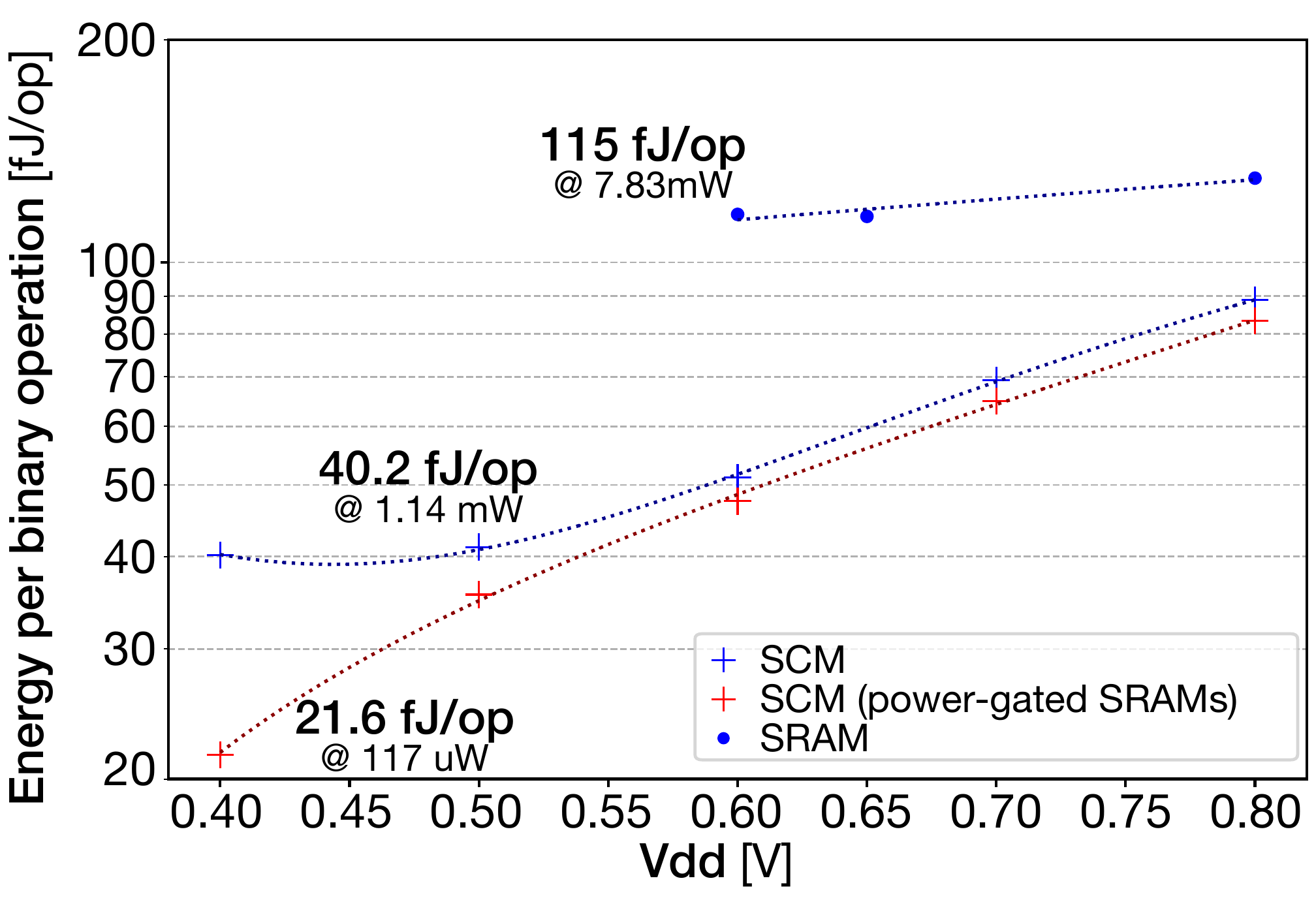}
    }
    \caption{Sustained performance and \uC{} system level energy efficiency, when using the XNE to execute binary convolutions on data stored in SRAM or SCM memories. Dotted lines are used for curve fitting between characterized operating points (crosses / circles).}
\end{figure*}

\begin{figure}[tb]
\footnotesize
\centering
    \includegraphics[clip, width=0.49\textwidth]{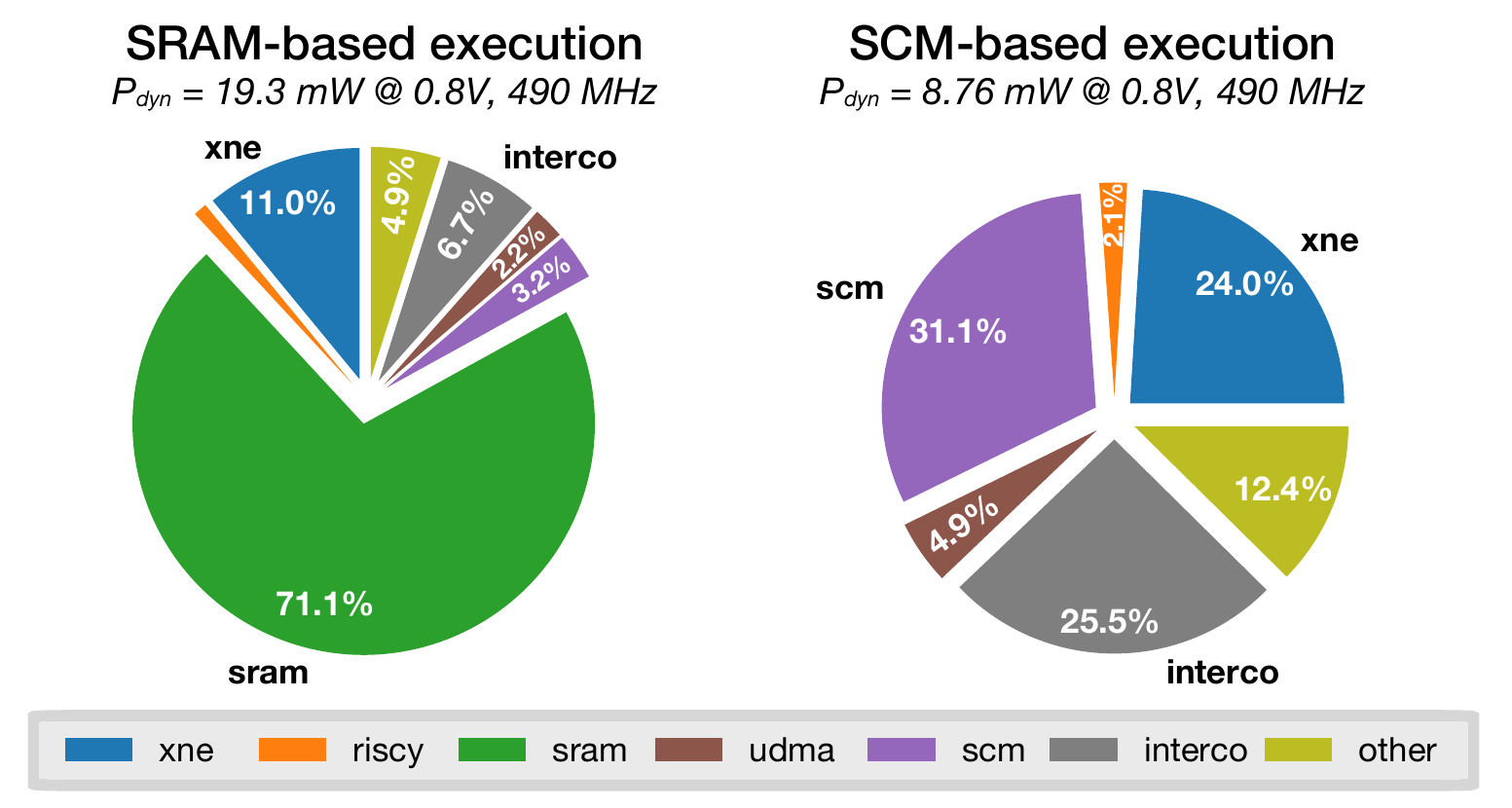}
    \caption{Distribution of dynamic power, when using the XNE to execute binary convolutions on data stored in SRAM or SCM memories.}
    \label{fig:dyn_pow_hier}
\end{figure}

The XNE is designed as a tightly-coupled accelerator engine~\cite{conti_ultralowenergy_2015} and it can be more completely evaluated when integrated within a full system-on-chip.
To this end, given the results shown in Section~\ref{sec:results_sweep}, we selected the design with \TP{}$=128$ for integration in a HW-accelerated microcontroller system (\uC{}).
The \uC{} uses the RISCY~\cite{GautschiNearThresholdRISCVCore2017} RISC-V ISA core and features also an autonomous I/O subsystem (uDMA)~\cite{PulliniuDMAautonomoussubsystem2017}, capable of moving data from/to memory and to/from selected I/O peripherals (SPI, I2C, I2S, UART, CPI, and HyperRAM) - and also of marshaling data in the shared memory\footnotemark.
\footnotetext{
	The MCU is based on a modified version of \textsc{PULPissimo} (\texttt{github.com/pulp-platform/pulpissimo}), which includes RISCY, uDMA and an example accelerator.
}
We targeted the 22nm technology referred in Section \ref{sec:results_sweep}; we used the same tools reported in Section~\ref{sec:results_sweep} for synthesis and backend.

Figure~\ref{fig:quentin_archi} shows the architecture of the \uC{} system and its floorplan, where the most relevant blocks have been highlighted.
The \uC{} is internally synchronous and memories, core and accelerator belong to a single clock domain.
The \uC{} has 64~kB of core-coupled memory accessed prioritarily by RISCY and 456 kB of memory shared between RISCY, uDMA and XNE.
Both kinds of memory are hybrids of SRAM banks and latch-based standard-cell-memory~\cite{AndriYodaNNArchitectureUltraLow2017} (SCM).
Specifically, 8~kB of core-coupled memory are made of multi-ported SCMs and 8~kB of shared memory are single-ported SCMs.
As will be detailed in the following of this section, SCMs are essential to keep the \uC{} operational below the rated operating voltage for SRAM memories, and they are also typically more energy-efficient than SRAMs, although they are much less area-efficient.
Finally, all SRAMs operate on a separate power domain and can be completely turned off by an external DC-DC converter.

\subsubsection{Performance evaluation}
To evaluate the performance of the XNE, we compare with an efficient software implementation targeted at low-power microcontrollers~\cite{RusciAlwaysONVisualnode}.
A naive implementation of the binary convolution kernel requires on average 2 cycles per each \textit{xnor-popcount}, which is clearly highly inefficient due to the extremely fine granularity of the operation.
By performing multiple convolutions on adjacent pixels in a parallel fashion, and the RISCY instructions for popcount, throughput can be increased by $\sim9\times$ up to 3.1 op/cycle\footnote{Throughout the paper, we count xnor and popcount as separate operations, therefore 1 xnor + 1 popcount = 2 op}.

On the other hand, the XNE integrated in the \uC{} system can sustain a throughput of 220 op/cycle under normal conditions (86\% of its theoretical peak throughput with \TP{}=128, with the drop being caused by memory contention and small control bubbles).
This means that the XNE can provide a net improvement of 71$\times$ to throughput for binary convolutions and densely connected layers with respect to optimized software.
Figure~\ref{fig:perf} shows the overall sustained throughput at the \uC{} system level in various operating points in typical conditions, with operating frequency extracted from PrimeTime timing analysis.
At the nominal operating point (\SI{0.8}{\volt}), the \uC{} works at up to \SI{490}{\mega\hertz} and the XNE can reach a throughput of up to 108 Gop/s.

\begin{figure*}[tb]
    \centering
    \subfloat[\textit{mVGG-d} topology based on Courbariaux~et~al.~\cite{CourbariauxBinarizedNeuralNetworks2016}. \label{fig:mini_vgg9}]{
        \includegraphics[clip, width=0.5\textwidth]{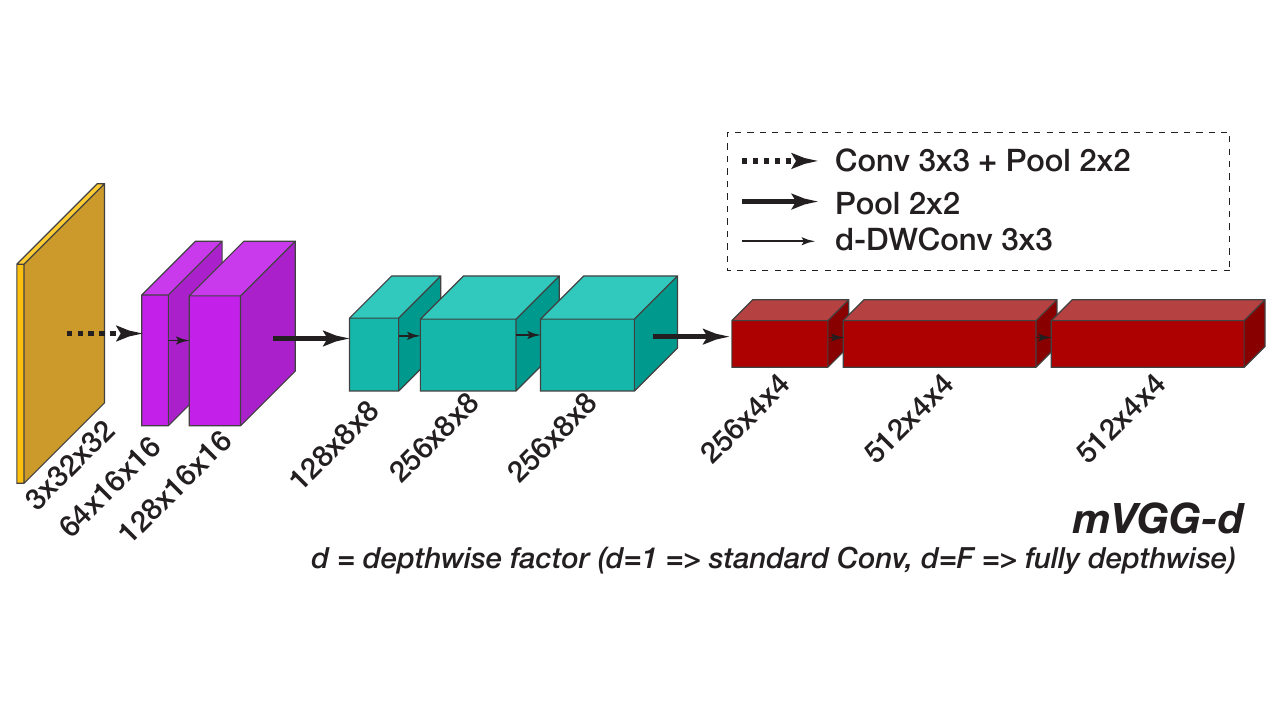}
    }
    \qquad
    \subfloat[Energy per inference vs Error. \label{fig:cifar10}]{
        \includegraphics[clip, width=0.4\textwidth]{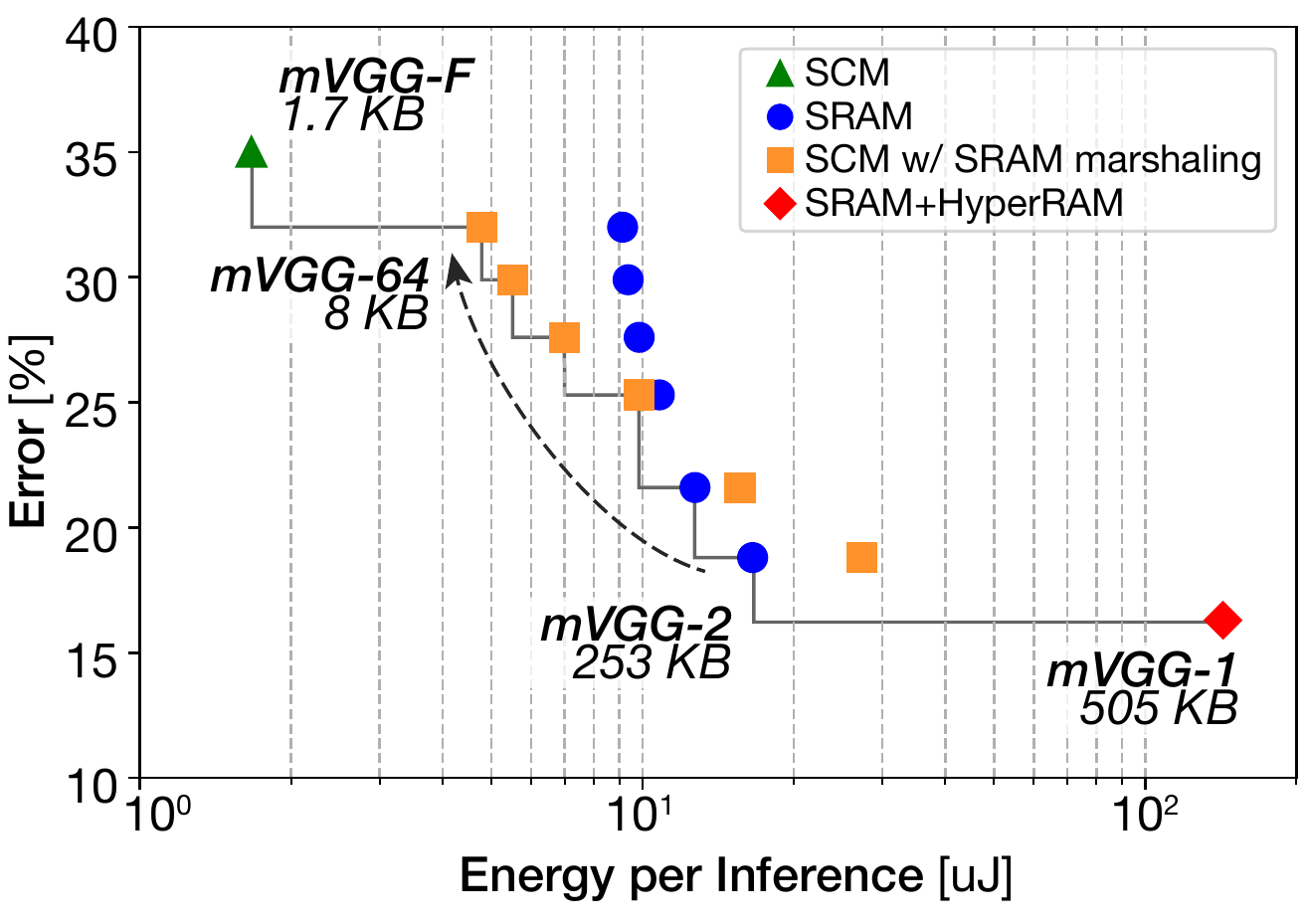}
    }
    \caption{mVGG binary neural network energy per inference vs error trade-off on \textit{mVGG-D}; in the rightmost plot, green triangles, blue circles, orange squares and red diamonds represent respectively usage modes on pure SCM @0.4V, on SRAM / on SCM with SRAM marshaling @0.6V, and with HyperRAM marshaling @0.6V. The grey solid line indicates the Pareto frontier.}
    \label{fig:cifar10_minivgg}
\end{figure*}

\subsubsection{Energy efficiency evaluation}
We evaluated separately the power consumption of the XNE when insisting on the SRAMs, which are rated for  operation between \SI{0.6}{\volt} and \SI{0.8}{\volt}, and on the SCMs, which we evaluated down to \SI{0.4}{\volt}.
Since SRAMs can be entirely switched-off externally, and the \uC{} does not depend on them for essential operations, we evaluated both the case in which they are fully switched off and the one in which they are simply not used (and therefore they consume static leakage power).

%
Figure \ref{fig:dyn_pow_hier} shows the outcome of this evaluation in terms of dynamic power at \SI{0.8}{\volt}, while executing an XNE-based binary convolution kernel either on data located on SRAM or on SCM.
When executing on the SRAM, the dynamic power due to memory clearly dominates over computation, by a factor of $7.1\times$, taking into account also the power spent in the system interconnect.
Conversely, SCM-based execution is more balanced, as SCMs consume $\sim$3$\times$ less then SRAMs.
In both cases, memory accesses are largely due to weights, which are loaded many times and used only once in the XNE design.

The advantage of working on SCMs is clearer when we evaluate energy efficiency in terms of femtoJoules per operation, as shown in Figure~\ref{fig:efficiency}.
There is a factor of $\sim$ 2-3$\times$ between SRAM- and SCM-based execution, especially when the operating voltage is reduced\footnote{According to the SRAM model we  used, the internal power which dominates in SRAMs is less dependent on Vdd than the net switching power which dominates in most other modules -- this is also the reason for which the energy efficiency in SRAM mode is flatter in Figure~\ref{fig:efficiency}.}.
SCMs, which are $\sim$2-3$\times$ less power-hungry and do not stop working at low voltage, enable the XNE to deliver much better energy efficiency.
If we do not fully switch down the SRAMs, the minimum energy point is located near the \SI{0.5}{\volt} operating point, where the \uC{} delivers 28 Gop/s and  \SI{40.2}{\femto\joule} per operation are required - equivalent to a system-level efficiency of 25 Top/s/W.
Power-gating the SRAMs vastly reduces leakage power and moves the minimum energy point further down in operating voltage: at \SI{\bestefficiency{}}{\femto\joule} per operation at \SI{0.4}{\volt}.

\subsubsection{Energy-accuracy tradeoff in BNNs}
\label{sec:results_energy_accuracy}

The most efficient use case for the \uC{} platform is clearly when entire network topologies can be fully deployed on the shared memory, and in particular on the SCM.
To fully showcase the impact of the model memory footprint on the overall efficiency, we used a simple topology derived from a reduced version of the popular VGG~\cite{SimonyanVeryDeepConvolutional2014}, as proposed by Courbariaux~et~al.~\cite{CourbariauxBinarizedNeuralNetworks2016}; we trained it on the CIFAR-10 dataset for 150 epochs using their same binarization strategy, ADAM optimizer, and initial learning rate 0.005.
Figure~\ref{fig:mini_vgg9} shows the \textit{mVGG-d} network.
To scale the number of parameters stored in memory in a smooth fashion, we kept the network architecture of \textit{mVGG-d} fixed, but progressively modified the nature of convolutional layers from the standard definition of \ref{eq:conv_base} in the direction of depthwise separable convolutions~\cite{CholletXceptionDeepLearning2016} following the parameter \textit{d}.
Specifically, we modeled convolutions of the form
\begin{equation}
    \mathbf{y}(k_{out}) = \mathrm{bin}_{\pm1}\left( \sum_{k_{in}=d\cdot k_{out}}^{(d+1)\cdot k_{out}-1} \Big( \mathbf{W}(k_{out},k_{in}) \otimes \mathbf{x}(k_{in}) \Big)\right)
    \label{eq:conv_depthwise}
\end{equation}
This model is fully supported by the XNE with minor microcode modifications.

To model power consumption in the various versions of \textit{mVGG-d}, we consider several usage modes.
When the network (parameters and partial results) fully fits within the shared SCM memory, we operate at the most efficient energy point -- \SI{0.4}{\volt} with power-gated SRAMs, consuming \SI{\bestefficiency{}}{\femto\joule} per operation. 
Conversely, when it does not fit the SCMs but fits in the SRAMs, we operate at \SI{0.6}{\volt}, consuming \SI{115}{\femto\joule} per operation.
As an alternative, we also support a mode in which weights, which are responsible for the majority of the energy consumption, are marshaled from SRAM to a temporary SCM-based buffer.
In this case, the energy cost of computation is reduced to \SI{52}{\femto\joule}, but there is an overhead of $\sim$\SI{8.7}{\pico\joule} per bit to move weights from SRAM to SCM. 
Finally, when the SRAM is too small to host the weights, they are stored in an external memory and loaded to the SRAM when needed by means of the uDMA.
In this case, we considering using a low-power Cypress HyperRAM 8MB DRAM memory~\cite{Cypress64Mbit128Mbit} as external memory, directly connected to the \uC{} uDMA.
The HyperRAM operates at \SI{125}{MHz} (1~Gbit/s) and \SI{28.6}{\pico\joule} per bit read.

Figure \ref{fig:cifar10} shows the results of this evaluation in terms of the Pareto plot of the size/energy versus accuracy trade-off in \textit{mVGG-d} BNNs. We scale $d$ with power-of-two values from 1 to 64 and consider also the case of fully depthwise separable convolutions (\textit{mVGG-F}).
The results clearly show the impact of memory energy on even small benchmarks such as \textit{mVGG-d}.
The most accurate model, \textit{mVGG-1}, is only $\sim$6\% from the current state-of-the-art for BNNs on CIFAR-10~\cite{CourbariauxBinarizedNeuralNetworks2016}; however, this model consumes roughly 10x of \textit{mVGG-2}, because it cannot run at all without the external HyperRAM.
Increasing \textit{d}, we observe that the energy penalty of marshaling data from SRAMs to SCMs is increasingly reduced up to a point (\textit{mVGG-8}) where it becomes less significant than the cost of operating directly on the SRAMs; hence it becomes convenient to marshal data between the two.
Finally, the \textit{mVGG-F} model is so small that it can be run entirely on SCMs and consumes 100$\times$ less than \textit{mVGG-d}, but it suffers a significany penalty in terms of accuracy.

\begingroup
\begin{table*}[tb]
\footnotesize
\centering
    \begin{tabulary}{\textwidth}{l C C C C C C C}
        \multirow{2}{*}{\textbf{Name}} & \multirow{2}{*}{\textbf{Technology}} & \multirow{2}{*}{\textbf{Maturity}} & \textbf{Core Area} & \textbf{Peak Perf.} & \textbf{Energy Eff.} & \textbf{On-chip Mem.} \\
         & & & [mm${}^2$] & [Top/s] & [Top/s/W] & [kB] \\
        \toprule
        \textit{BRein}~\cite{AndoBReinMemorySingleChip2017} & 65nm & silicon & 3.9 & 1.38 & 6 & - \\
        \textit{XNOR-POP}~\cite{JiangXNORPOPprocessinginmemoryarchitecture2017} & 32nm & layout & 2.24 & $\sim$5.7 & $\sim$24 & 512 \\
        \textit{UNPU}~\cite{LeeUNPU506TOPS} & 65nm & silicon & 16 & 7.37 & 51 & 256 \\
        \textit{XNORBIN}~\cite{BahouXNORBIN95TOp2018} & 65nm & layout & 1.04 & 0.75 & 95 & 54 \\
        Bankman~et~al.~\cite{BankmanAlwaysOn8mJ86} & 28nm mixed-signal & silicon & 4.84 & - & 722  & 329 \\
        \midrule
        \textbf{This work (\uC{}, SCM w/ SRAM off)} & \textbf{22nm} & \textbf{layout} & \textbf{2.32} & \textbf{0.11} & \textbf{46} & \textbf{520} \\
        \textbf{This work (\uC{}, SCM)} & \textbf{22nm} & \textbf{layout} & \textbf{2.32} & \textbf{0.11} & \textbf{25} & \textbf{520} \\
        \textbf{This work (XNE \TP{}=128)} & \textbf{22nm} & \textbf{-} & \textbf{0.016} & \textbf{0.11} & \textbf{112} & \textbf{-} \\
        \textbf{This work (XNE \TP{}=128)} & \textbf{65nm} & \textbf{-} & \textbf{0.092} & \textbf{0.07} & \textbf{52} & \textbf{-} \\
        
        \bottomrule
    \end{tabulary}
    \caption{Comparison of Hardware Accelerators and Application-Specific ICs for Binary Neural Networks}
    \label{tab:comparison}
\end{table*}
\endgroup


\subsubsection{Real-world sized BNN execution}
\label{sec:results_resnet}
The size of real-world state-of-the-art DNN topologies for most interesting problems is such that it does not make sense at all to consider fully localized execution on the 520~kB of on-chip memory of the \uC{} system, even with BNNs.
Supporting execution aided by external platforms is, therefore, critical.
To minimize the continuous cost that would be implied by transfer of partial results, we dimensioned the \uC{} system so that relatively big BNN topologies can be run using the external memory exclusively for storing weights.

As representatives of real-world sized BNNs, we chose ResNet-18 and ResNet-34~\cite{HeDeepResidualLearning2015}, which can be fully binarized providing a top-5 accuracy of 67.6\% and 76.5\% respectively on the ImageNet database~\cite{LinAccurateBinaryConvolutional2017}.
A binarized implementation of the ResNets requires \SI{128}{\kilo\byte} for input, output and partial results buffering (taking into account also shortcut connections), plus a maximum of \SI{288}{\kilo\byte} for the weights of a single layer; the final densely connected layer requires more memory, but it has an extremely small footprint for partial result buffering, and therefore it is possible to efficiently divide the computation in filtering tiles executed sequentially.
Overall, it is possible to execute both these topologies on the tiny XNE-equipped \uC{} system without any energy cost for moving partial results.

To evaluate how efficient the deployment of such a model can be, we consider the same system of Section~\ref{sec:results_energy_accuracy}, with an 8 MB HyperRAM connected to the uDMA.
We consider the SRAM-based execution mode for this evaluation.
We consider weights to be transferred asynchronously by means of the uDMA, performing double buffering to overlap memory access by the XNE with the fetching of the next set of weights.
ResNet-18 and ResNet-34 require $3.64\times 10^9$ and $7.34 \times 10^9$ operations respectively.
In this operating mode, the compute time dominates for all layers except the last group of convolutions and the final fully connected layer in both ResNet-18 and ResNet-34.
ResNet-18 inference can be run at $\sim$14.7 fps, spending \SI{1.45}{\milli\joule} per frame on a standard 224$\times$224 input; for the latter at 8.9 fps, spending \SI{2.17}{\milli\joule} per frame.

In both cases, the contribution of memory traffic to energy consumption is dominant, mostly due the final layers (especially the fully connected one, which is memory-bound).
The impact of these layers is more relevant in ResNet-18 than in ResNet-34, hence memory traffic energy is more dominant in the former case (by 2.5$\times$) than in the latter (by 60\%).
Even if the cost of memory traffic cannot be entirely removed, the design of the \uC{} system mitigates this cost by making most data movements unnecessary, as weights are directly loaded on the shared SRAM and partial results never have to leave it.

\subsubsection{Comparison with the state-of-the-art and discussion}
\label{sec:results_comparison}
Table \ref{tab:comparison} shows a comparison between our work and the current state-of-the-art in hardware accelerators for Binary Neural Networks.
Contrary to our solution, current systems do not implement a full microcontroller or System-on-Chip, but consist either in near-memory computing techniques (\textit{BRein}, \textit{XNOR-POP}) or dedicated ASICs for binary neural networks.

Of all the ASIC accelerators taken into account, Bankman~et~al.~\cite{BankmanAlwaysOn8mJ86} claims by far the highest energy efficiency (more than 700 Top/s/W), but they are dependent on full-custom mixed signal IPs that are known to be delicate with respect to on-chip variability and difficult to port between technologies.
Moreover, their approach has hardwired convolution size (2$\times$2), which severely limits their flexibility to implement different kinds of convolutions.

\textit{XNORBIN}~\cite{BahouXNORBIN95TOp2018} achieves the second-best result with a much more traditional fully-digital ASIC architecture, achieving almost 100 Top/s/W with a 65nm chip.
Compared with our \uC{} design, the main advantage of \textit{XNORBIN} is placed in its custom memory hierarchy, enabling a non-constrained design for what concerns the accelerator core.
This fact accounts for most of its advantage in terms of raw energy efficiency.
However, \textit{XNORBIN} does not include enough memory to implement BNNs bigger than AlexNet and, in general, it does not have facilities to enable exchange of data with the external world.
Similarly, \textit{UNPU}~\cite{LeeUNPU506TOPS} targets efficient execution without particular attention to communication.
It is roughly 16$\times$ bigger than XNORBIN, but reaches only half the energy efficiency.

Compared to \textit{UNPU} and \textit{XNORBIN}, the best fully digital designs currently in the state-of-the-art (to the best of our knowledge), our work tackles a different problem: not providing the lowest energy solution as-is, but a methodology and an accelerator IP for the integration of BNNs within a more complete System-on-Chip solution, with an eye to system level problems, in particular the cost of memory accesses.
The XNE has been designed to \replaced{make efficient use of the relatively limited memory bandwidth allowed in an MCU-like SoC}{use its available memory bandwidth efficiently} (the interfaces are active $\sim$95\% of the overall execution time in many cases) and to be small and unobtrusive in terms of area ($\sim$1.5\% of the proposed \uC{}) and timing closure (30\% shorter critical path than the overall \uC{} system).
\added{Conversely, the design of an ASIC accelerator deals with different architectural constraints -- in particular, the memory hierarchy is designed around the accelerator to provide the maximum effective memory bandwidth.
For example, \textit{XNORBIN} uses an ad-hoc memory hierarchy in which weights, feature maps and lines are stored separately (the datapath is fed by a linebuffer) – amounting for improved effective memory bandwidth available with respect to our design (and hence higher efficiency), at the expense of flexibility and of area.
}
\deleted{
Moreover, the peak performance provided by the XNE in the proposed \uC{} system is well-matched with the sustainable external memory bandwidth in a microcontroller, as argued in Sections~\ref{sec:results_energy_accuracy} and \ref{sec:results_resnet}; many of the works in the state-of-the-art target irrealistic peak performance with respect to their available external bandwidth.
}

\replaced{
To the best of our knowledge, the XNE-accelerated \uC{} is the only design that can execute \textit{software-defined} BNNs in an efficient way, by taking advantage of the tight integration between the XNE accelerator, the RISCY core and the uDMA to speed up nested loops of binary matrix-vector products.
The generality of this mechanism makes the \uC{} capable of dealing with all BNNs in which the linear part of convolutional and fully connected layers is constituted of binary matrix-vector products (a group which contains most known neural network topologies), provided that the external memory can store all weights.
}
{
Finally, to the best of our knowledge, the XNE-accelerated \uC{} is the only design capable of efficiently executing \textit{arbitrary} BNNs  thanks to the tight integration between the XNE accelerator, the RISCY core and the uDMA -- as demonstrated in Section~\ref{sec:results_resnet} for what concerns ResNet-18 and ResNet-34.
This makes it significantly more flexible than most of the current state-of-the-art in BNN acceleration.
}



\section{Conclusion}
\label{sec:conclusion}
To the best of our knowledge, this paper is the first to introduce a fully synthesizable ultra-efficient hardware accelerator IP for binary neural networks meant for integration within microcontroller systems.
We also propose a microcontroller system (\uC{}) designed to be flexible and usable in many application scenarios, but at the same time extremely efficient (up to 46 Top/s/W) for BNNs. The \uC{} is the only work in the current state-of-the-art capable of executing real-world sized BNN topologies such as ResNet-18 and ResNet-34; the latter can be run in \SI{2.2}{\milli\joule} per frame in real time (8.9 fps).
As a third contribution, we also performed an analysis of the relative costs of computation and memory accesses for BNNs, showing how the usage of a hardware accelerator can be significantly empowered by the availability of a hybrid memory scheme.

A prototype based on the \uC{} system presented in Section~\ref{sec:results_uC} has been taped out in 22nm technology at the beginning of January 2018.
Future work includes silicon measurements on the fabricated prototype; the extension of this design to explicitly target more advanced binary neural network approaches, such as ABC-Net~\cite{LinAccurateBinaryConvolutional2017}; and as more advanced integration with the SRAM memory system to reduce power in high-performance modes and enable more parallel access from the accelerator while keeping the shared memory approach. 

\bibliographystyle{IEEEtran}

\newpage

\begin{IEEEbiography}[{\includegraphics[width=0.85in,clip,keepaspectratio]{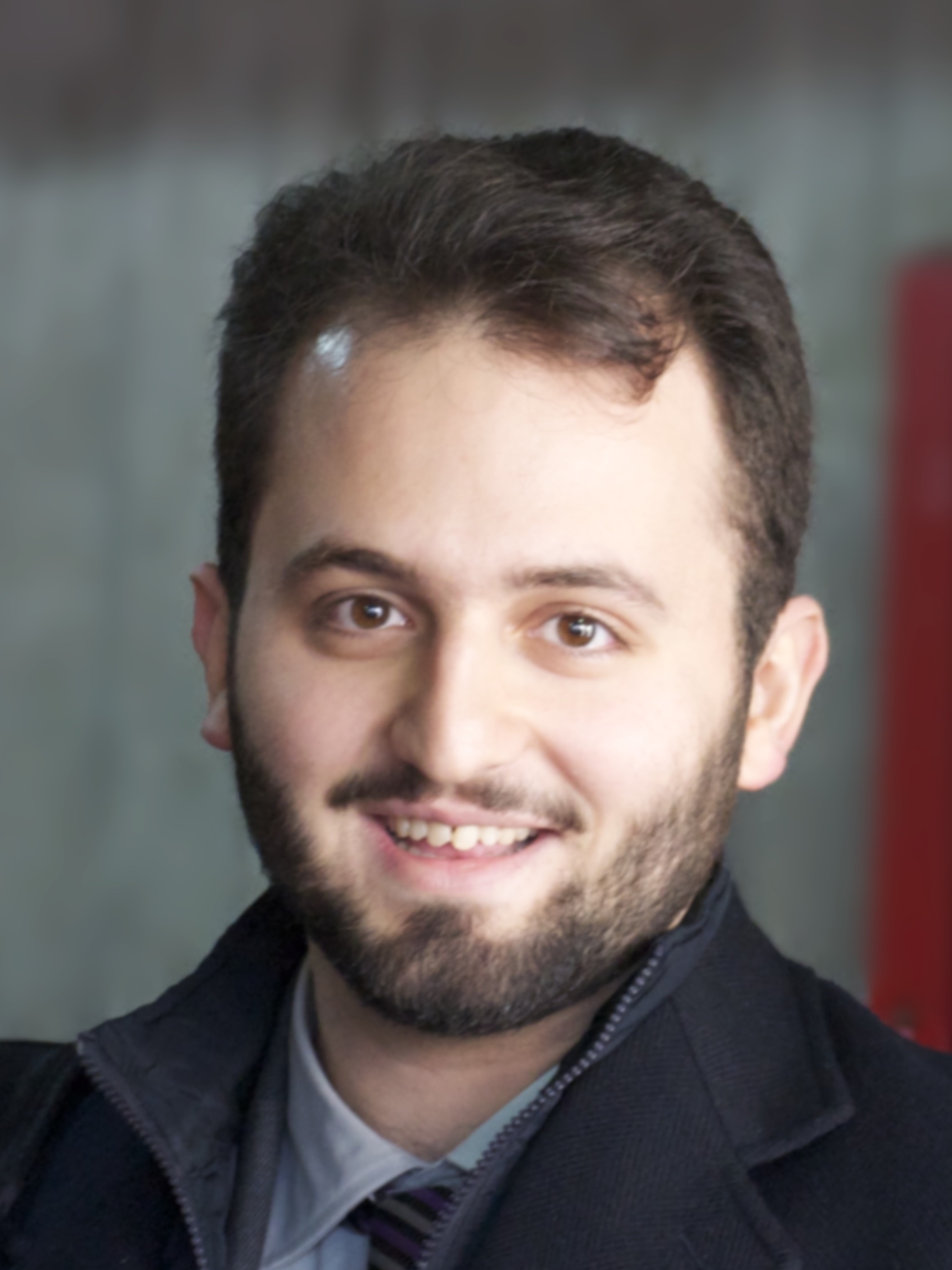}}]{Francesco Conti} received the Ph.D. degree from University of Bologna in 2016 and is currently a post-doctoral researcher at the Integrated Systems Laboratory, ETH Z\"{u}rich, Switzerland and the Energy-Efficient Embedded Systems laboratory, University of Bologna, Italy.
He has co-authored more than 20 papers on international conferences and journals.
His research focuses on energy-efficient multicore architectures and applications of deep learning to low power digital systems.
\end{IEEEbiography}

\begin{IEEEbiography}[{\includegraphics[width=0.85in,clip,keepaspectratio]{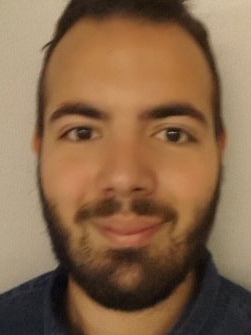}}]{Pasquale Davide Schiavone} received his B.Sc. (2013) and M.Sc. (2016) in computer engineering from Polytechnic of Turin. Since 2016 he has started his PhD studies at the Integrated Systems Laboratory, ETH Zurich. His research interests include low-power microprocessors design in multi-core systems and deep-learning architectures for energy-efficient systems.
\end{IEEEbiography}

\begin{IEEEbiography}[{\includegraphics[width=0.85in,clip,keepaspectratio]{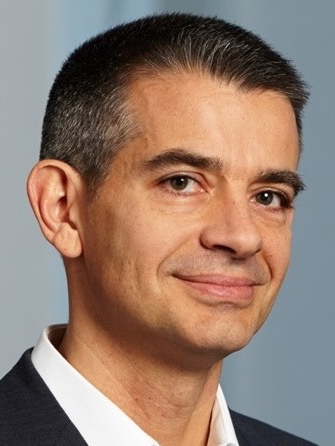}}]{Luca Benini} holds the chair of Digital Circuits and Systems at ETH Z\"{u}rich and is Full Professor at the Universit\`{a} di Bologna.
Dr. Benini's research interests are in energy-efficient system design for embedded and high-performance computing. 
He has published more than 800 papers, five books and several book chapters.
He is a Fellow of the ACM and a member of the Academia Europaea. He is the recipient of the 2016 IEEE CAS Mac Van Valkenburg award.
\end{IEEEbiography}

\vfill






\end{document}